\documentclass{article}

\usepackage{microtype}
\usepackage{graphicx}
\usepackage{subcaption}
\usepackage{booktabs}
\usepackage{hyperref}
\usepackage{tcolorbox}
\tcbuselibrary{breakable}
\usepackage{multirow}

\usepackage[accepted]{eiml_icml2026}

\usepackage{amsmath}
\usepackage{amssymb}
\usepackage{mathtools}
\usepackage{amsthm}
\usepackage{parskip}

\usepackage[capitalize,noabbrev]{cleveref}

\theoremstyle{plain}
\newtheorem{theorem}{Theorem}[section]

\theoremstyle{definition}

\newtheorem{assumption}[theorem]{Assumption}
\theoremstyle{remark}

\usepackage[textsize=tiny]{todonotes}

\icmltitlerunning{Rethinking Uncertainty Evaluation in Large Language Models}

\begin{document}

\twocolumn[
  \icmltitle{Rethinking Uncertainty Evaluation in Large Language Models}

  \icmlsetsymbol{equal}{*}

  \begin{icmlauthorlist}
    \icmlauthor{Krish Matta }{equal,yyy}
    \icmlauthor{Atharv Naphade }{equal,yyy}
    \icmlauthor{Andy Zou }{meta}
  \end{icmlauthorlist}

  \icmlaffiliation{yyy}{Carnegie Mellon University}
  \icmlaffiliation{meta}{Meta}
  \icmlcorrespondingauthor{Krish Matta}{self@krishmatta.net}

  \icmlkeywords{Machine Learning, ICML, Confidence Estimation, Calibration, Large Language Models, Uncertainty Quantification, Epistemology, confidence calibration, self-consistency, trustworthy ML}

  \vskip 0.3in
]
%% ============================================================
%% REPLACE your tcolorbox preamble block with exactly this:
%% (requires \usepackage[most]{tcolorbox} and \usepackage{xcolor})
%% ============================================================

%% ============================================================
%% PREAMBLE: replace your entire tcolorbox block with this.
%% Requires: \usepackage[most]{tcolorbox}  and  \usepackage{xcolor}
%% ============================================================

%% ============================================================
%% USAGE IN BODY: wrap each group of same-color boxes like this,
%% with \vspace{2pt} between consecutive boxes in the same group,
%% and \smallskip before/after the whole group:
%%
%%   \smallskip
%%   \begin{tcolorbox}[structural] ... \end{tcolorbox}
%%   \vspace{2pt}
%%   \begin{tcolorbox}[structural] ... \end{tcolorbox}
%%   \vspace{2pt}
%%   \begin{tcolorbox}[structural] ... \end{tcolorbox}
%%   \smallskip
%%

%% Also add this just before your first tcolorbox group in the .tex body,
%% and after each group ends, to suppress parskip inflation around boxes:
%%
%%   \setlength{\parskip}{0pt}   % <- before first box in a group
%%   ... boxes ...
%%   \setlength{\parskip}{6pt}   % <- restore after group (match your doc's parskip)
%%
%% ============================================================

\printAffiliationsAndNotice{~}

%% ============================================================
%% ============================================================
%% TIGHTENED ABSTRACT (replace existing abstract with this)
%% ============================================================
%% ============================================================
%% TIGHTENED ABSTRACT (replace existing abstract with this)
%% ============================================================
 %% ============================================================
%% TIGHTENED ABSTRACT (replace existing abstract with this)
%% ============================================================

%% ============================================================
%% TIGHTENED INTRODUCTION (replace existing introduction)
%% ============================================================
\begin{abstract}
Calibration is the primary criterion for evaluating LLM confidence, but it is insufficient: it admits trivially incoherent estimators, depends on the evaluation distribution, and does not test the extent to which the estimation can be interpreted as a consistent, underlying probability function. What we actually need is for LLM confidence estimates to satisfy the conditions required of coherent probabilistic beliefs. We formalize these conditions along three axes (structural coherence, faithfulness, and usefulness) and operationalize them as the $\mathbf{C1}$ metrics. Widely used estimators systematically violate these conditions despite appearing well-calibrated: models assign lower confidence to logically easier questions 31\% of the time, and common interventions reducing RMSCE leave structural violations unchanged, suggesting that calibration is orthogonal to probabilistic validity. RLHF and chain-of-thought improve usefulness metrics without restoring coherence. Our results show current LLM confidence estimates cannot be interpreted as coherent probabilities; our framework provides the tools to measure and close this gap.
 
\end{abstract}
 
%% ============================================================
%% INTRODUCTION
%% ============================================================
\section{Introduction}
\label{sec:introduction}
 
Reliable confidence estimates in LLMs would support principled abstention, cascading to stronger models, and uncertainty-aware aggregation in agentic pipelines. The field has centered on \emph{calibration} as the primary criterion \cite{geng2024surveyconfidenceestimationcalibration, phan2025humanitysexam, openai2024gpt4technicalreport}, but calibration alone is inadequate. A constant predictor that always outputs the model's overall accuracy is perfectly calibrated yet carries no instance-level information. Moreover, an estimator calibrated on a benchmark can be arbitrarily miscalibrated on sub-populations. Most fundamentally, calibration ignores internal coherence: a model assigning high confidence to mutually exclusive answers, or deeming a hard question easier than one it logically implies, violates probability axioms while passing calibration tests.
\begin{figure}
    \centering
    \includegraphics[width=1.0\linewidth]{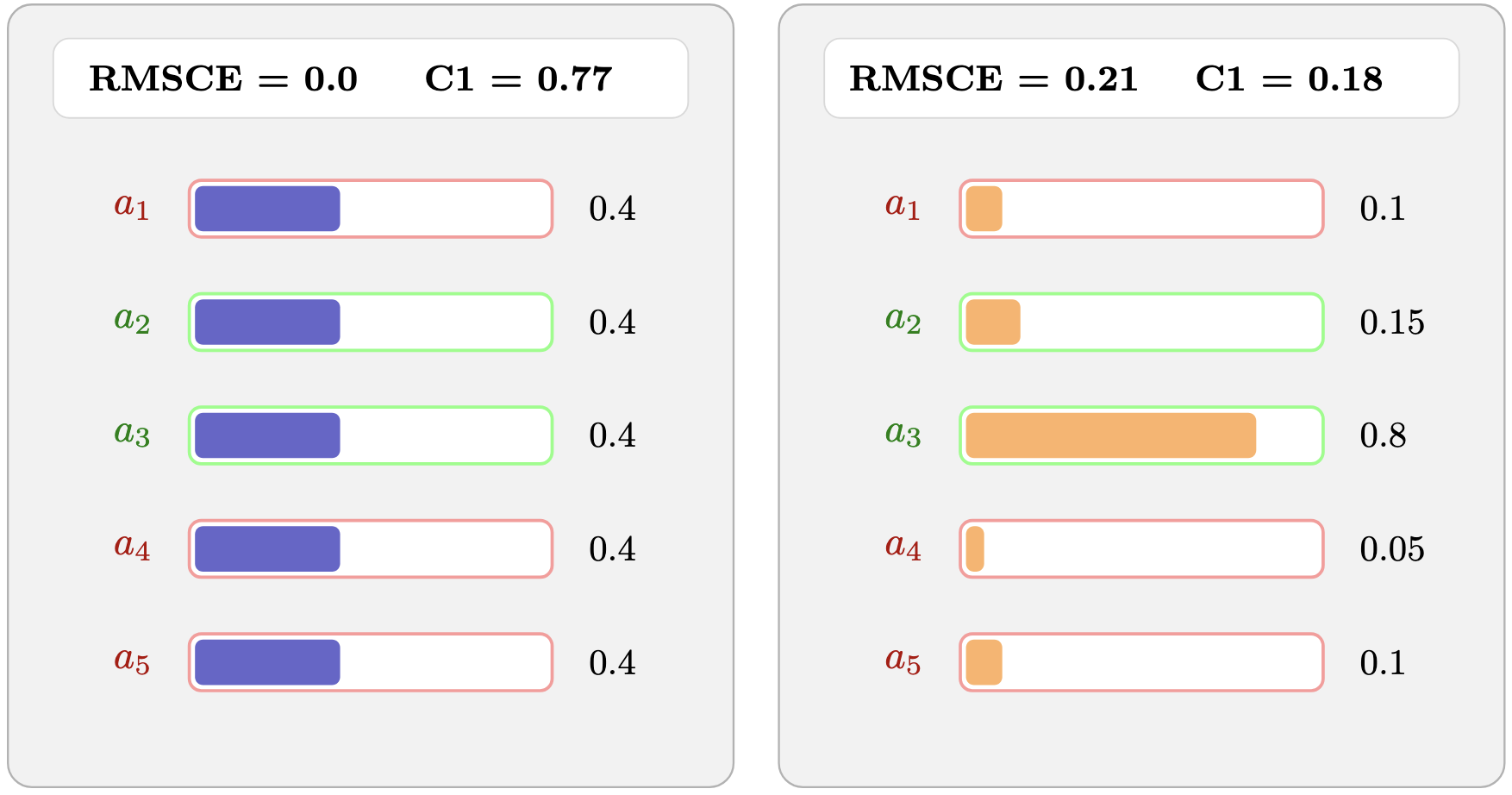}
    \caption{C1 Metrics reduce the flaws in Calibration}
    \label{fig:placeholder}
\end{figure}
We propose a richer framework grounded in rational belief theory \cite{ramsey1926, Cox1946-COXPFA-3} and the utility engineering approach of \cite{mazeika2025utilityengineeringanalyzingcontrolling}, defining three categories: \textbf{structural properties} (normalization, conjunction consistency, entailment monotonicity); \textbf{faithfulness properties} (prompt and generation semantic invariance); and \textbf{usefulness properties} (calibration, discrimination). We instantiate this framework as \textbf{C1} and evaluate verbal \cite{tian2023justaskcalibrationstrategies}, logit-based \cite{kadavath2022languagemodelsmostlyknow}, and SliCK \cite{gekhman2024doesfinetuningllmsnew} confidence estimators. Output-based estimators saturate near certainty, masking structural failures; SliCK is the only estimator with meaningful calibration (RMSCE 0.251 vs.\ 0.778, 0.700) and discrimination (AUROC 0.825 vs.\ 0.559, 0.596), while exposing that the underlying model violates entailment monotonicity 31\% of the time. % Scaling analysis across LLaMA-3 reveals that usefulness improves with scale while prompt semantic invariance worsens monotonically---larger models become more sensitive to surface rephrasing, a failure mode orthogonal to calibration.
 
\textbf{Contributions.}
\begin{enumerate}
    \item We show that calibration is insufficient: it admits incoherent and irrational confidence estimators in practice.
\item We introduce structural coherence, faithfulness, and usefulness, and show that standard estimators achieve apparent calibration while masking severe probability violations. 
\item Using interventions, we diagnose how training and inference differences impact performance across our metrics. 
\end{enumerate}

%% ============================================================
%% ============================================================
\section{Related Work}
\label{sec:related}
%% ============================================================

\textbf{Calibration as a Standard.}
Calibration is the de facto standard for evaluating LLM confidence. Recent surveys \cite{geng2024surveyconfidenceestimationcalibration, zhou2024surveylargelanguagemodels}, benchmarks like Humanity's Last Exam \cite{phan2025humanitysexam}, and technical reports \cite{openai2024gpt4technicalreport} center on metrics like RMSCE. 

\textbf{Output-Based Confidence.}
Output-based estimators directly utilize token log-probabilities \cite{kadavath2022languagemodelsmostlyknow, ye2024assessingcreativityllmsproposing} or explicitly prompted verbalized scores \cite{lin2022teachingmodelsexpressuncertainty, tian2023justaskcalibrationstrategies}. Though verbalized confidence often achieves lower calibration error \cite{tian2023justaskcalibrationstrategies}, it suffers from prompt sensitivity \cite{xiong2024llmsexpressuncertaintyempirical} and domain-specific overconfidence \cite{phan2025humanitysexam, zhou2024surveylargelanguagemodels}. We evaluate both as baselines and find that the strong calibration reported in prior work does not hold under our setup; their score saturation both inflates apparent structural consistency and produces severe miscalibrations on harder factual QA.

\textbf{Consistency and Sampling.}
Alternatively, sampling-based methods measure output consistency, positing that correct answers are generated more stably than hallucinations \cite{wang2023selfconsistencyimproveschainthought, manakul2023selfcheckgptzeroresourceblackboxhallucination}. This is formalized by SliCK \cite{gekhman2024doesfinetuningllmsnew}, which calculates sample agreement rates, and Semantic Entropy \cite{kuhn2023semanticuncertaintylinguisticinvariances}, which clusters by meaning instead of exact tokens. While computationally expensive, these methods give richer uncertainty estimates. We show this exposes structural failures hidden by output-based methods.

\textbf{Coherence and Consistency.}
While prior work explores consistency (e.g., against prompt paraphrasing \cite{elazar2021measuringimprovingconsistencypretrained}), it generally treats it as a performance metric rather than a probabilistic requirement. Closest to our framework is the utility engineering approach of \cite{mazeika2025utilityengineeringanalyzingcontrolling}, which audits whether LLM preferences satisfy rational utility axioms. We apply similar logic to confidence: structural coherence is a necessary condition for model outputs to be reliable and predictable.

\section{Motivation \& Formalization}

\subsection{Confidence Functions and Estimators}
\label{sec:confidence-functions}

Let $\mathcal{M}$ be a language model, $x$ a prompt, and $y$ a candidate
response. Let $\simeq$ denote a semantic equivalence relation over strings,
and write $[x]$ and $[y]$ for the corresponding equivalence classes. For a
prompt class $[x]$, let
$\mathcal{Y}_x := \{[y] : y \text{ is a candidate answer to } x\}$
denote the set of semantically distinct answer classes. We assume
correctness is defined at the level of equivalence classes, and that all
semantically correct answers belong to a single gold class
$[y^\star] \in \mathcal{Y}_x$.\footnote{When multiple surface forms are
candidates, they are grouped into a single equivalence class via $\simeq$.}

\paragraph{Beliefs.}
Following the subjectivist tradition \cite{ramsey1926,Cox1946-COXPFA-3}, we
model the model's epistemic state as a probability measure $P$ over a space
$\Omega$ of world-states.\footnote{This is the same measure $P$ underlying
the structural axioms in Appendix~\ref{app:theory}.} Each prompt class
induces a random variable
\[
A_{[x]} : \Omega \to \mathcal{Y}_x,
\]
\[
A_{[x]}(\omega) = \text{the answer class correct for } [x]
\text{ in world } \omega ,
\]
whose randomness is purely epistemic: the correct answer is fixed, but the
model is uncertain which world it inhabits. The model's
\emph{confidence function} is its credence that a given class is correct,
\[
c([x],[y]) := P\bigl(A_{[x]} = [y]\bigr).
\]
By construction, $c([x],\cdot)$ is the distribution of $A_{[x]}$ under
$P$, hence a probability distribution over $\mathcal{Y}_x$. In
particular, since exactly one answer class is correct in each world, the
events $\{A_{[x]} = [y]\}_{[y] \in \mathcal{Y}_x}$ partition $\Omega$ and
$\sum_{[y]} c([x],[y]) = 1$. Moreover, as $c$ is defined on equivalence classes, paraphrased
answers receive identical credence by construction.

This definition admits a direct decision-theoretic reading: under 0--1
loss, a rational agent outputs
$\hat{y}([x]) := \arg\max_{[y]} c([x],[y])$, and its self-assessed
probability of answering correctly is the \emph{prompt-level confidence}
\[
\bar{c}([x]) := \max_{[y]} c([x],[y]) = P\bigl(A_{[x]} = \hat{y}([x])\bigr).
\]
The probabilistic interpretation is justified by standard coherence
arguments: under the Dutch book argument \cite{ramsey1926}, $c([x],[y])$ is
the fair price of a contract paying \$1 if $[y]$ is correct, and Cox's
theorem \cite{Cox1946-COXPFA-3} implies that any consistent system of
beliefs over such events is isomorphic to a probability measure.
Crucially, because all prompts induce events over the \emph{same} space
$\Omega$, credences on logically related prompts (e.g.\ a multi-hop
question and its sub-questions) are marginals and conditionals of a single
measure, motivating the structural properties in
Section~\ref{sec:structural-properties}.

We emphasize that $c$ is a normative reference object: we do not assume that the model's actual epistemic state admits such a representation. Whether its observed confidence behavior is consistent with \emph{any} coherent credence function is precisely what our metrics evaluate.

\paragraph{Generation.}
The credence $c$ is latent: it cannot be read off the model's weights.
What we can observe is the model's \emph{generation distribution}.
Sampling a response to $x$ at temperature $T$ induces a random variable
$Z_x \in \mathcal{Y}_x$, the semantic class of the sampled response, with
distribution
\[
g([x],[y]) := P^{\mathrm{gen}}(Z_x = [y]).
\]
Unlike $A_{[x]}$, the randomness in $Z_x$ is aleatory. It arises from
stochastic decoding, not from uncertainty about the world, hence $g$ is not
equivalent to $c$ by definition. Connecting the two requires an explicit assumption:

\begin{assumption}[Sampling faithfulness]
\label{ass:faithfulness}
At the evaluation temperature, the model's generation distribution over
semantic answer classes coincides with its credence:
$g([x],\cdot) = c([x],\cdot)$ for all $[x]$.
\end{assumption}

This assumption is the (usually tacit) operating premise of the
self-consistency literature \cite{wang2023selfconsistencyimproveschainthought,kuhn2023semanticuncertaintylinguisticinvariances,
gekhman2024doesfinetuningllmsnew}: output variability under sampling is taken to reflect
epistemic uncertainty.

\paragraph{Estimators.}
We evaluate computable \emph{confidence estimators}
$\hat{c} : (x, y) \mapsto [0,1]$ that approximate $c([x],[y])$ from text,
differing in which route to $c$ they take:
\begin{itemize}
\item \textbf{Output-based} methods elicit a self-report of $c$ conditioned
on a single generation: verbalized scores (verbal) or normalized
verification logits (logit-based). These require no faithfulness
assumption, but presuppose accurate introspection and calibrated
verbalization.
\item \textbf{Sampling-based} methods (e.g.\ SliCK) estimate $g$ by Monte
Carlo over $k$ i.i.d.\ rollouts,
\[
\hat{c}(x,y) \approx \frac{\lvert\{j : y_j \in [y]\}\rvert}{k},
\]
which is consistent for $g$ unconditionally, and consistent for $c$ under
Assumption~\ref{ass:faithfulness}.
\end{itemize}
The distinction between confidence \emph{functions} and confidence \emph{estimators} is important: structural properties
(Section~\ref{sec:structural-properties}) are defined over the full distribution
$c([x],\cdot)$, and therefore require estimators that meaningfully
approximate it. A structural violation measured via a sampling-based
estimator thus has two possible sources: an incoherent credence $c$, or a
failure of Assumption~\ref{ass:faithfulness}.
Either source is disqualifying for downstream uses that require a coherent probabilistic interpretation. We further develop these foundations in
Appendix~\ref{app:theory}.

% \paragraph{Estimators.}
% The true confidence function $c$ is not directly observable. Instead, we
% evaluate \emph{confidence estimators} $\hat{c}$, computable functions:
% \[
% \hat{c} : (x,y) \mapsto [0,1],
% \]
% which approximate $c([x],[y])$ from text.

% Different estimators provide different approximations:
% \begin{itemize}
% \item Output-based methods (verbal, logit-based) estimate confidence
% conditioned on a single generation.
% \item Sampling-based methods (e.g., SliCK) approximate the full distribution
% over equivalence classes via Monte Carlo sampling:
% \[
% \hat{c}(x,y) \approx \frac{|\{j : y_j \in [y]\}|}{k},
% \]
% which is a consistent estimator of $c([x],[y])$ under i.i.d.\ sampling.
% \end{itemize}

% This distinction is central: structural properties (e.g., normalization,
% entailment) are defined over the full distribution $c([x],\cdot)$, and
% therefore require estimators that meaningfully approximate it.
% We further clarify and prove all claims regarding the confidence function and benchmark validity in Appendix~\ref{app:metric-clarification}.

\subsection{Why Calibration Is Insufficient}
\label{sec:calibration-insufficient}

Calibration is the primary evaluation criterion for confidence estimation in the LLM literature. A confidence function $c$ is \emph{calibrated} on a distribution of prompt-generation pairs $\mathcal{D}$ if for all $p \in [0, 1]$, $P_{(x, y) \sim \mathcal{D}}([y]\text{ is correct for } [x] \mid c([x], [y]) = p) = p$. Calibration is typically evaluated using the root-mean-square calibration error (RMSCE) \cite{phan2025humanitysexam}, which bins confidence scores into $B = 20$ equal-width bins over $[0, 1]$ and computes:
\[
\text{RMSCE} = \sqrt{\sum_{b=1}^{B} \frac{n_b}{N} \left(\text{acc}_b - \mu_b\right)^2}
\]
where $n_b$ is the number of pairs in bin $b$, $\text{acc}_b$ is the fraction correct, and $\mu_b$ is the mean confidence. Surveys of LLM uncertainty quantification \cite{geng2024surveyconfidenceestimationcalibration} organize the field around calibration as the primary desideratum, and recent benchmarks such as Humanity's Last Exam \cite{phan2025humanitysexam} report calibration error as a central evaluation metric. The GPT-4 technical report's finding that RLHF degrades calibration relative to the base model \cite{openai2024gpt4technicalreport} has further entrenched it as the canonical evaluation metric for confidence.

Unfortunately, calibration as a sole criterion admits confidence functions that are internally incoherent. We illustrate this with two examples.

\textbf{Constant predictor.} Let $\alpha$ denote the model's overall accuracy on $\mathcal{D}$. The confidence function $c([x], [y]) = \alpha$ for all $(x, y)$ is perfectly calibrated on $\mathcal{D}$, achieving RMSCE of exactly zero. Since $c$ is constant, every prediction falls into the single bin $b^{\star}$ containing $\alpha$, so $\mu^{\star} = \alpha = \text{acc}_{b^\star}$ and the sum vanishes:
\[
\text{RMSCE} = \sqrt{\frac{N}{N}(\alpha - \alpha)^2} = 0.
\]
Yet $c$ carries no instance-level information and cannot distinguish a question the model answers reliably from one it answers by chance.

\textbf{Calibration is distribution-relative.} More fundamentally, calibration is not a property of $c$ alone, but of $c$ paired with $\mathcal{D}$. A function well-calibrated on $\mathcal{D}$ can be arbitrarily miscalibrated on sub-distributions of $\mathcal{D}$ itself. 

Let $\mathcal{D}' \subset \mathcal{D}$ be the sub-distribution of examples the model answers incorrectly. On $\mathcal{D}'$, every bin has accuracy zero, so each bin contributes a strictly positive term to the squared RMSCE:
$$(n'_b / N') \cdot (\bar{c}'_b)^2$$
Hence:
$$\text{RMSCE}(\mathcal{D}') > 0 = \text{RMSCE}(\mathcal{D})$$
Any partition of $\mathcal{D}$—by topic, difficulty, or domain—yields sub-distributions on which $c$ may be substantially miscalibrated. RMSCE on a benchmark thus characterizes aggregate behavior and provides no guarantee about the sub-populations that matter in deployment.
%% ============================================================
%% ============================================================
\section{Methodology}
\label{sec:methodology}
%% ============================================================
In this section we describe our multidimensional evaluation of Confidence, \textbf{C1}. % We can take the average of resulting sub-metrics as a single C1 score.
\subsection{Axioms for Confidence Evaluation}
\label{sec:framework}

Drawing on the decision-theoretic foundations of rational belief (the Dutch book argument \cite{ramsey1926}; Cox's theorem \cite{Cox1946-COXPFA-3}), we define three orthogonal categories for evaluating confidence in language models. \textbf{Structural properties} are hard constraints derived from the probability axioms: whether reported confidences normalize, respect the product rule, and respect logical entailment. \textbf{Faithfulness properties} ask whether a confidence estimator faithfully represents the underlying confidence function, requiring invariance to surface-level rephrasing. \textbf{Usefulness properties} ask whether confidence tracks ground truth, encompassing calibration and discrimination. 

% Throughout, we assume the model is a logically consistent actor whose beliefs constitute a single probability measure $P$ over world-states $\Omega$; each question $x$ induces a correctness event $E_x \subseteq \Omega$, and all confidence statements are marginals or conditionals of the same $P$. See Appendix~\ref{app:theory} for full justification.

\subsubsection{Structural Properties}
\label{sec:structural-properties}
Structural metrics test rationalizability: whether the confidences an
estimator reports can be explained by any single probability measure
(Section~\ref{sec:confidence-functions}). A violation refutes the
conjunction of two hypotheses: that the model's underlying beliefs are
coherent, and that the estimator faithfully reflects them. The
faithfulness metrics (Section~\ref{sec:faithfulness-properties}) target the second
hypothesis specifically, enabling partial attribution: when an estimator
satisfies the faithfulness properties yet exhibits structural
violations, the incoherence is attributable to the model's beliefs rather than the measurement.

% Structural properties are hard constraints on $c$ derived from the probability axioms. Violations indicate that the model's beliefs are internally contradictory.

\textbf{Normalization} requires that confidences over the answer classes
sum to one: the classes partition the response space, so the marginals
of any probability measure satisfy $\sum_{[y]} c([x],[y]) = 1$, and the
credence $c$ satisfies this by construction
(Section~\ref{sec:confidence-functions}). The metric tests whether
\emph{reported} confidences can be rationalized by any such credence.
We measure the normalization deviation $|S(x)-1|$ on 1,500 SimpleQA
\cite{wei2024measuringshortformfactualitylarge} questions with $k=16$ rollouts per question at
temperature $T=0.5$, where
\[
S(x) = \sum_j \frac{1}{|[y_j]|} \sum_{y \in [y_j]} \hat{c}(x, y)
\]
averages confidence within each equivalence class to avoid
double-counting surface forms of the same answer.

\textbf{Conjunction Consistency} requires that if correctly answering $[x]$ decomposes into a first-hop sub-question $[x_1]$ with gold answer $[y_1^*]$ followed by a second-hop sub-question $[x_2]$, then
\[
\bar{c}([x]) = c(x_1, y^*_1) \cdot \bar{c}([x_2]\mid [x_1],[y_1^*]),
\]
by the product rule $P(A\cap B)=P(A)\cdot P(B\mid A)$. We measure the deviation
\[
\Delta(x)=\bigl|\bar{\hat{c}}(x)-\hat{c}(x_1, y^*_1)\cdot\bar{\hat{c}}(x_2\mid x_1,y_1^*)\bigr|
\]
on 2-hop MuSiQue \cite{trivedi2022musaboringmultihop} questions ($k=16$, $T=0.5$), with $\bar{\hat{c}}$ estimated as max confidence across rollouts.

\textbf{Entailment Monotonicity.} Suppose answering $[x]$ correctly
entails answering $[x']$ correctly, i.e.
\[
\begin{aligned}
&\{\omega \in \Omega : A_{[x]}(\omega) = \hat{y}([x])\} \\
\subseteq\; & \{\omega \in \Omega : A_{[x']}(\omega) = \hat{y}([x'])\}:
\end{aligned}
\]
every world in which the model's chosen answer to $[x]$ is correct is
one in which its chosen answer to $[x']$ is correct. Monotonicity of
probability measures then requires
\[
\bar{c}([x]) \le \bar{c}([x']),
\]
so reported confidences with
$\bar{\hat{c}}([x]) > \bar{\hat{c}}([x'])$ cannot be rationalized by any
single measure.

We instantiate the entailment via MuSiQue's two-hop structure: $x'$ is
the second-hop sub-question conditioned on the gold first-hop answer
$y_1^*$, so answering the full question entails answering $x'$, and the
inclusion holds by construction of the dataset. We measure the violation magnitude
\[
\Delta(x) = \max\bigl(0,\; \bar{\hat{c}}(x) - \bar{\hat{c}}(x_2 \mid x_1, y_1^*)\bigr)
\]
on MuSiQue \cite{trivedi2022musaboringmultihop} ($k=16$, $T=0.5$), with
$\bar{\hat{c}}$ estimated as max confidence across rollouts.

\subsubsection{Faithfulness Properties}
\label{sec:faithfulness-properties}

Faithfulness properties constrain $\hat{c}$ to be consistent with a well-formed underlying $c$. Since $c$ is defined over equivalence classes, a faithful estimator must be invariant to surface-level reformulation. Violations indicate that $\hat{c}$ is sensitive to features of the text that are invisible at the equivalence class level, and therefore cannot faithfully represent $c$.

\textbf{Prompt Semantic Invariance} requires that for semantically equivalent prompts $x \simeq x'$:
\[
\hat{c}(x,y) = \hat{c}(x',y).
\]
We measure the deviation $\Delta(f)=|\bar{\hat{c}}(x)-\bar{\hat{c}}(x')|$ across paraphrase pairs from ParaRel \cite{elazar2021measuringimprovingconsistencypretrained}, sampling 1,500 facts with two paraphrase templates each ($k=16$, $T=0.5$).

\textbf{Generation Semantic Invariance} requires that for semantically equivalent generations $y \simeq y'$:
\[
\hat{c}(x,y) = \hat{c}(x,y').
\]
We measure the within-class spread $\Delta([y])=\max_{y\in[y]}\hat{c}(x,y)-\min_{y\in[y]}\hat{c}(x,y)$ across equivalence classes on 1,500 SimpleQA questions ($k=16$, $T=0.5$).

\subsubsection{Usefulness Properties}
\label{sec:usefulness-properties}

Usefulness properties ask whether confidence tracks ground truth. Unlike structural and faithfulness properties, they are distribution-relative by design.

\textbf{Calibration} requires that for all $p\in[0,1]$:
\[
P\bigl([y]\text{ correct}\mid c([x],[y])=p\bigr)=p.
\]
Among all pairs assigned confidence $p$, exactly a fraction $p$ should be correct. We measure RMSCE on 1,500 SimpleQA questions ($k=16$, $T=0.5$, $B=20$ bins); lower is better.

\textbf{Discrimination} requires that correct generations receive higher confidence than incorrect ones. This is distinct from calibration: a constant predictor achieves perfect calibration but chance-level discrimination. We measure AUROC over all (confidence, correctness) pairs on the same SimpleQA sample; higher is better.
\begin{figure}[t!]
\centering
\includegraphics[width=\columnwidth]{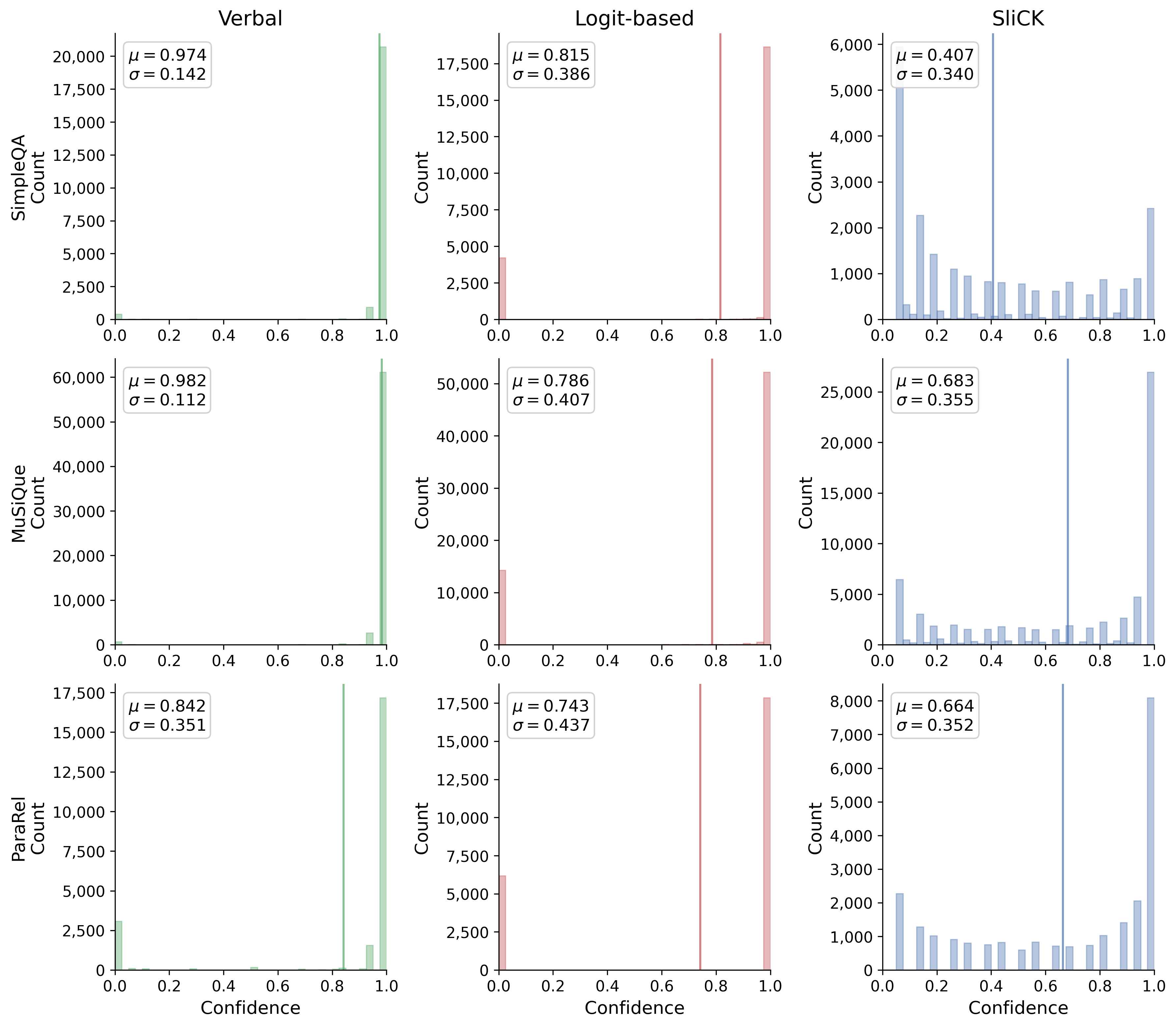}
\caption{\label{fig:confidence-histograms}Confidence score distributions across estimators (columns) and datasets (rows). Verbal and logit-based saturate at extremes; SliCK spans the full range.}
\end{figure}
\subsection{Confidence Estimation Methods}
\label{sec:methods}

We compare three representative estimators.

\textbf{Verbal Confidence} \cite{tian2023justaskcalibrationstrategies}. The model generates a response $y$ to $x$, then is asked in a follow-up to state the probability its answer is correct. The parsed numerical response is $\hat{c}(x,y)$.

\textbf{Logit-based Confidence} \cite{kadavath2022languagemodelsmostlyknow}. The model is prompted to verify whether $y$ is true or false for $x$; confidence is the normalized true-token probability:
\[
  \hat{c}(x,y) = \frac{P(\text{True})}{P(\text{True})+P(\text{False})}.
\]

\textbf{SliCK} \cite{gekhman2024doesfinetuningllmsnew}. We sample $k=16$ rollouts $y_1,\ldots,y_k$ to $x$, group them into equivalence classes under $\hat{\simeq}$ via LLM-as-a-judge, and exclude refusals and truncated outputs (letting $k'$ denote remaining rollouts). Confidence is the fraction of equivalent rollouts:
\[
  \hat{c}(x,y) = \frac{|\{j : y_j \mathrel{\hat{\simeq}} y\}|}{k'}.
\]

%% ============================================================
%% ============================================================
%% REPLACE existing \section{Experimental Results} with this version
%% ============================================================
\begin{figure}[t!]
\centering
\includegraphics[width=\columnwidth]{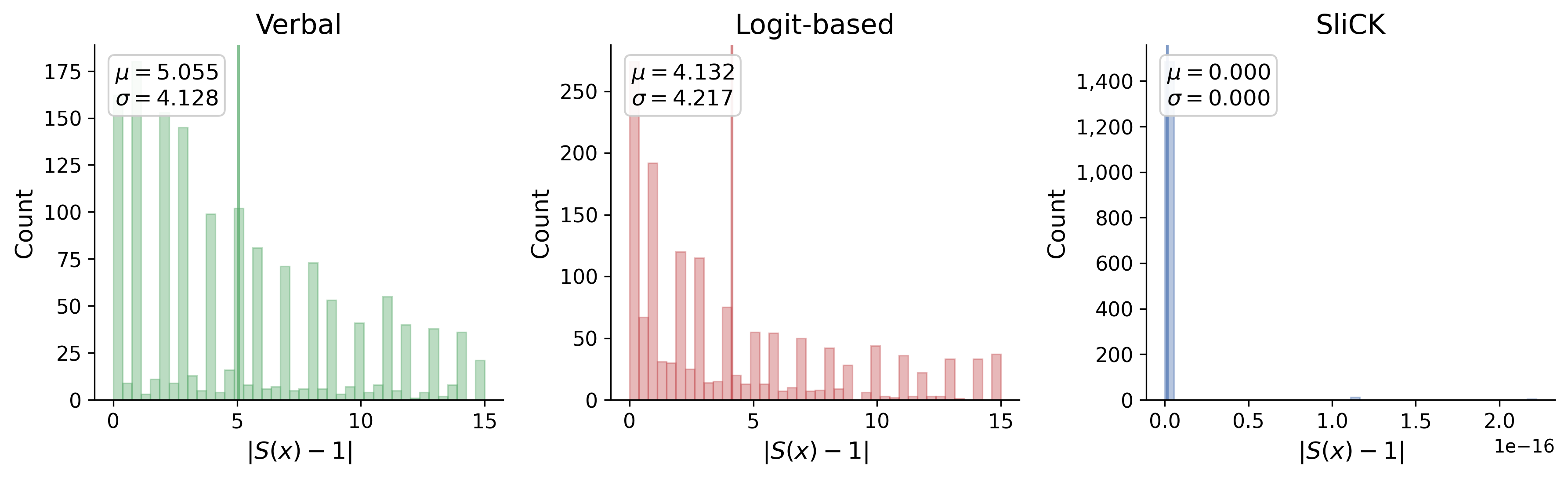}
\caption{\label{fig:normalization}Normalization deviation $|S(x)-1|$. Output-based estimators violate severely (5.055, 4.132); SliCK satisfies exactly by construction.}
\end{figure}

\begin{figure}[t!]
\centering
\includegraphics[width=\columnwidth]{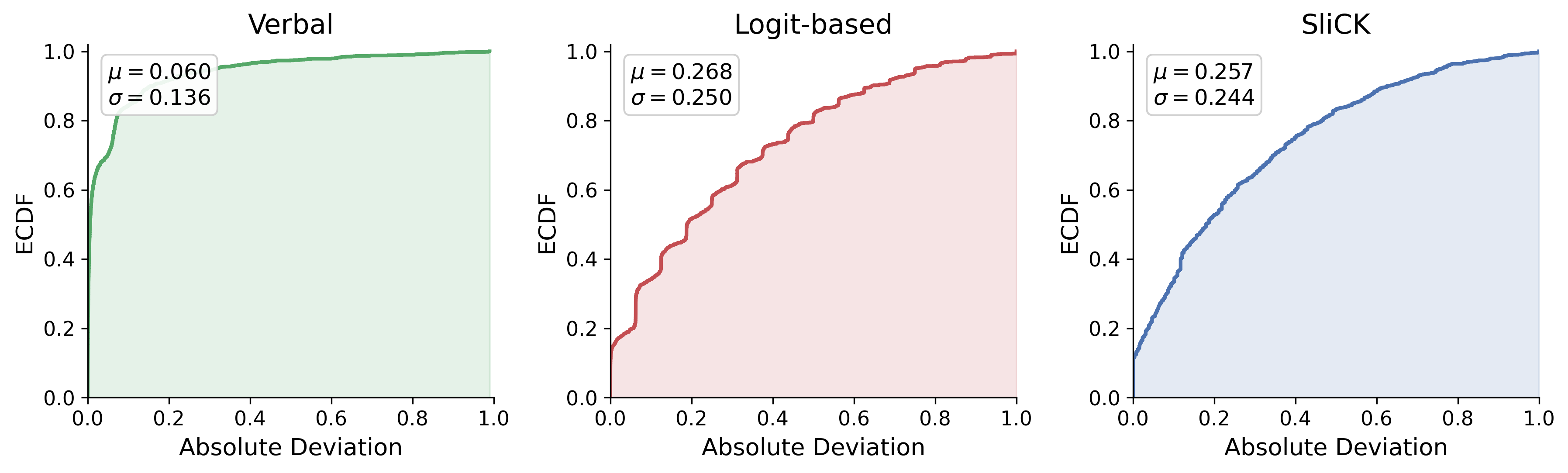}
\caption{\label{fig:conjunction}Conjunction consistency on MuSiQue. Verbal's low deviation is a saturation artifact; SliCK and logit-based reveal genuine violations.}
\end{figure}
%% ============================================================
%% REPLACE existing \section{Experimental Results} with this version
%% ============================================================

%% ============================================================
%% REPLACE existing \section{Experimental Results} with this version
%% ============================================================

The main experiments use \textbf{Qwen-30B-A3B-Thinking} \cite{yang2025qwen3technicalreport}, a 30B-parameter mixture-of-experts reasoning model with 3B active parameters, serving as both generation and evaluation model. Generations in which the model declines to answer or exhausts its token limit are excluded. Section~\ref{sec:scaling-and-reasoning} evaluates 9 additional models on a 200-question subset of each task with the same judge; full per-model results are in Appendix~\ref{app:results}, this ensures that results are not judge bias related.  

\subsection{What does C1 Reveal about Confidence Estimators?}
\label{sec:estimator-comparison}
By evaluating different estimators on C1, we identify failure modes of LLMs.
%We note upfront two design choices that shape interpretation. First, we evaluate raw logit-based and verbal confidence without post-hoc calibration (e.g., temperature scaling). This is intentional: post-hoc rescaling can reduce RMSCE but does not alter the structural outputs---a recalibrated estimator still assigns the same relative ordering and normalization behavior. We verify this in Appendix~\ref{app:results}. Second, SliCK satisfies normalization and generation semantic invariance \emph{by construction}; we include these not as empirical victories for SliCK but as diagnostic anchors that reveal how far output-based estimators deviate from provably achievable baselines. Three patterns dominate the remaining empirical results.

\paragraph{Saturation leads to poor C1}

Verbal and logit-based confidence concentrate nearly all mass at extreme values (Figure~\ref{fig:confidence-histograms}), and this directly corrupts their structural scores. We find that normalization deviation averages 5.055 (verbal) and 4.132 (logit-based). Their low scores on conjunction consistency (0.060), entailment monotonicity (4.9\% violations), and prompt invariance (0.025) are consequences of extreme values: $1.0 \approx 1.0 \times 1.0$ trivially. \textbf{Calibration} error alone \textbf{systematically rewards this} failure mode.

\begin{figure}[t!]
\centering
\includegraphics[width=\columnwidth]{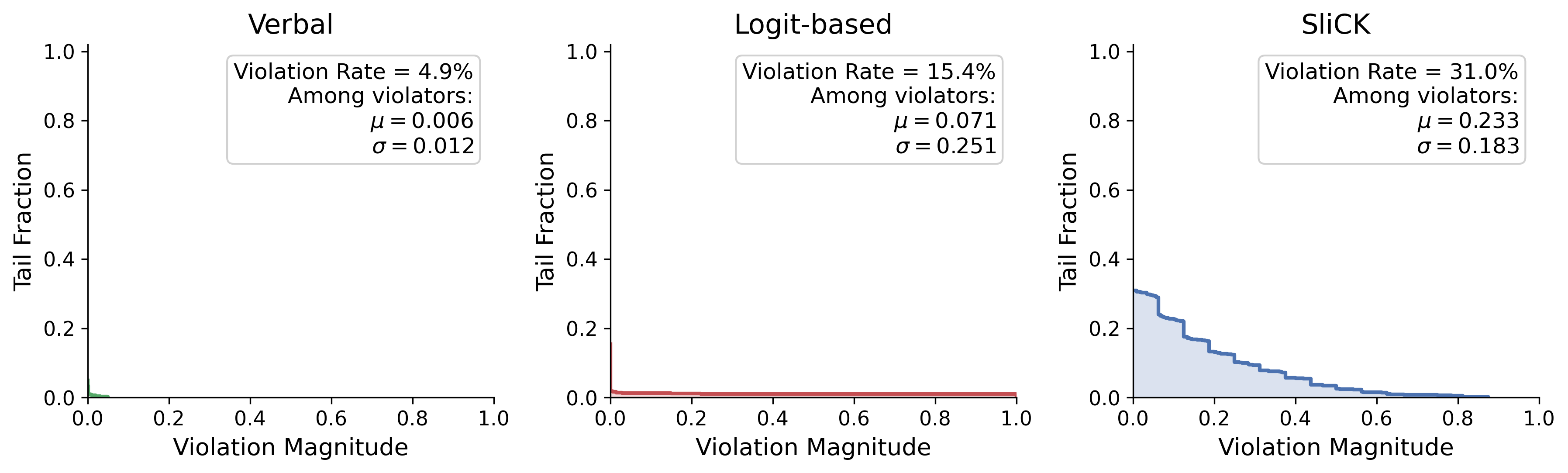}
\caption{\label{fig:entailment}Entailment monotonicity violations. SliCK: 31.0\% violation rate; apparent compliance of saturated estimators is a ceiling effect.}
\end{figure}

\begin{figure}[t!]
\centering
\includegraphics[width=\columnwidth]{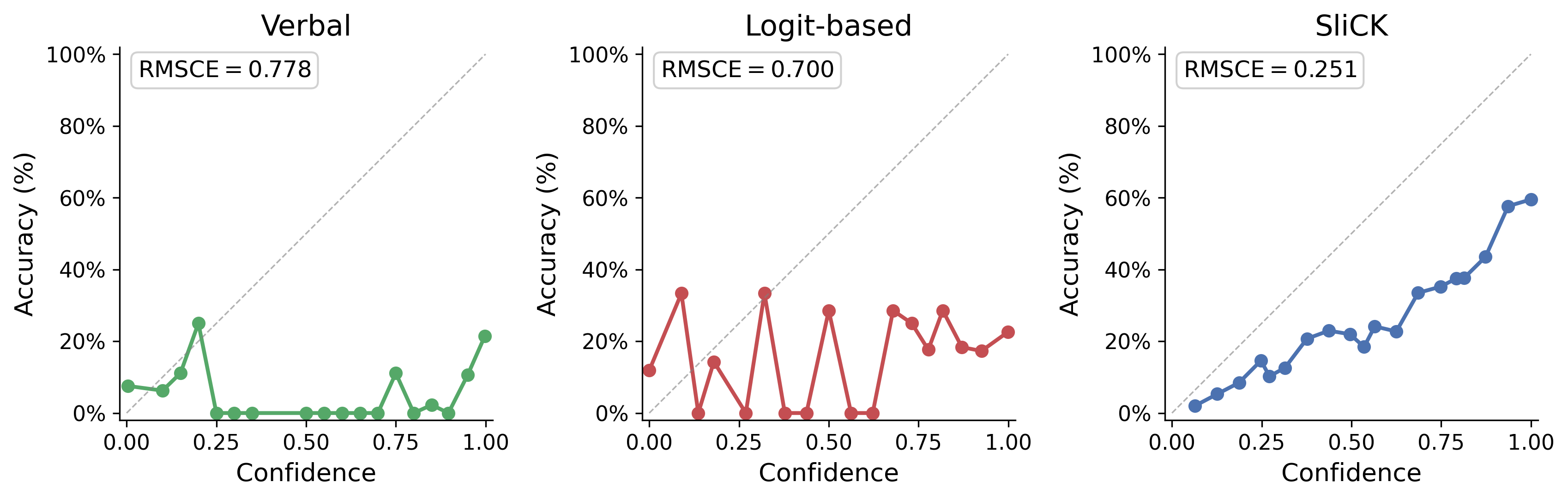}
\caption{\label{fig:calibration}Calibration diagrams and confidence distributions. SliCK alone tracks correctness; output-based estimators report near-certain confidence regardless of correctness.}
\end{figure}

\paragraph{SliCK surfaces genuine model incoherence.}
While SliCK achieves RMSCE 0.251 versus 0.778 and 0.700, and AUROC 0.825 versus 0.559 and 0.596 (Figure~\ref{fig:calibration}), SliCK exposes structural probability violations. Conjunction consistency deviation averages 0.257 on MuSiQue, comparable to logit-based (0.268), and entailment monotonicity is violated on \textbf{31.0\%} of questions (Figure~\ref{fig:conjunction},~\ref{fig:entailment}). Providing the first-hop gold answer makes the second hop strictly easier by construction, yet confidence decreases frequently and substantially. These are model-level failures that saturated estimators cannot surface.

\begin{figure}[t!]
\centering
\includegraphics[width=\columnwidth]{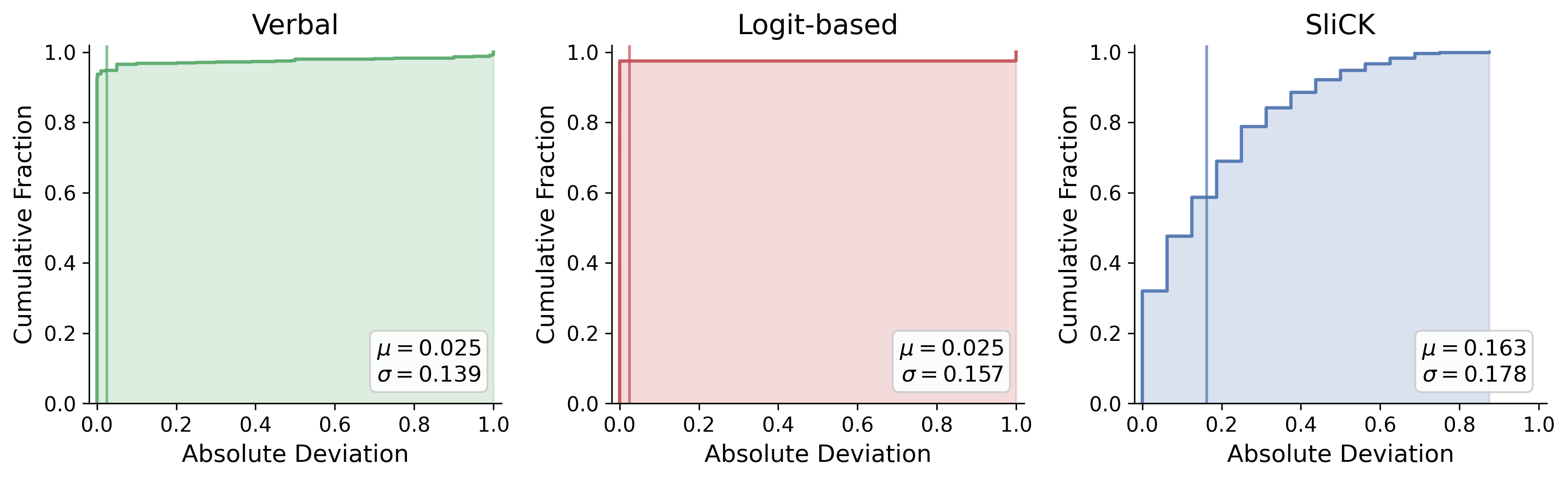}
\caption{\label{fig:prompt-invariance}Prompt semantic invariance on ParaRel. Output-based estimators vacuously consistent; SliCK genuinely sensitive ($\mu = 0.163$).}
\end{figure}

\begin{figure}[t!]
\centering
\includegraphics[width=\columnwidth]{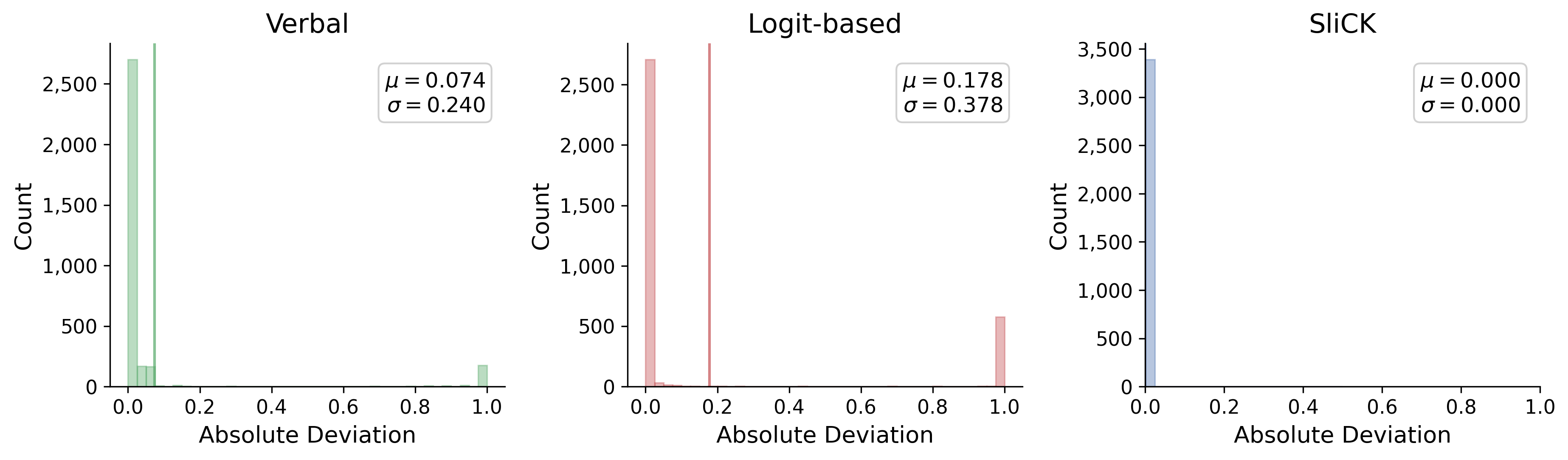}
\caption{\label{fig:generation-invariance}Within-class generation spread. SliCK exactly invariant; verbal and logit-based string-dependent.}
\end{figure}

\paragraph{Faithfulness failures differ by estimator}
SliCK trivially satisfies normalization, while verbal and logit-based estimators assign maximally different scores to generations they consider semantically equivalent (Figure~\ref{fig:generation-invariance}). SliCK's own failure is prompt invariance: paraphrases yield $\mu = 0.163$ deviation (Figure~\ref{fig:prompt-invariance}), a consequence of independent rollout sampling per prompt.

\subsection{How do factors impact C1 and Calibration?}
\label{sec:scaling-and-reasoning}

We re-run the evaluation on 200 samples per task (for computational limitations) across 9 models. The full results can be found in Appendix~\ref{app:results}, and we highlight notable findings. 

\paragraph{Model Size.} As shown in Figure~\ref{fig:scaling_laws}, Model size does not correlate cleanly with most coherence metrics, but Semantic Invariance shows a clear scaling trend across 9 models: smaller models are more sensitive to subtle prompt changes, with LLaMA-3-3B exhibiting 40\% more average n-gram diversity over 16 rollouts than LLaMA-3-70B.

% \paragraph{Model Size} 
 % While model size does not correlate, in some cases at all, with performance across Coherence Metrics, (see Appendix~\ref{app:results}) we find a clear scaling law for Semantic Invariance across 9 models. This suggests that smaller models are more drasticially effected by subtle prompt changes, evidenced by 40\% more average n-gram diversity over 16 rollouts on llama3-3B vs llama3-70b
 
% \paragraph{Impact of RLHF}
% Comparing Llama-3.1-70b with Nemotron-Llama-3.1-70b(graph in Appendix~\ref{app:results}, a model trained only via RLHF rewards \cite{wang2024helpsteer2preference}, RLHF modestly improves SliCK calibration (RMSCE: $0.365 \to 0.325$), it catastrophically collapses discriminability to chance (AUROC: $0.591 \to 0.498$). By uniformly inflating output confidence regardless of correctness, RLHF erases variance, leaving verbal calibration static and actively worsening conjunction consistency ($+15\%$). This reveals a critical vulnerability: modern alignment optimizes for confident outputs at the direct expense of accurate uncertainty quantification and structural coherence.

\begin{figure}[htbp]
    \centering
    \includegraphics[width=\linewidth]{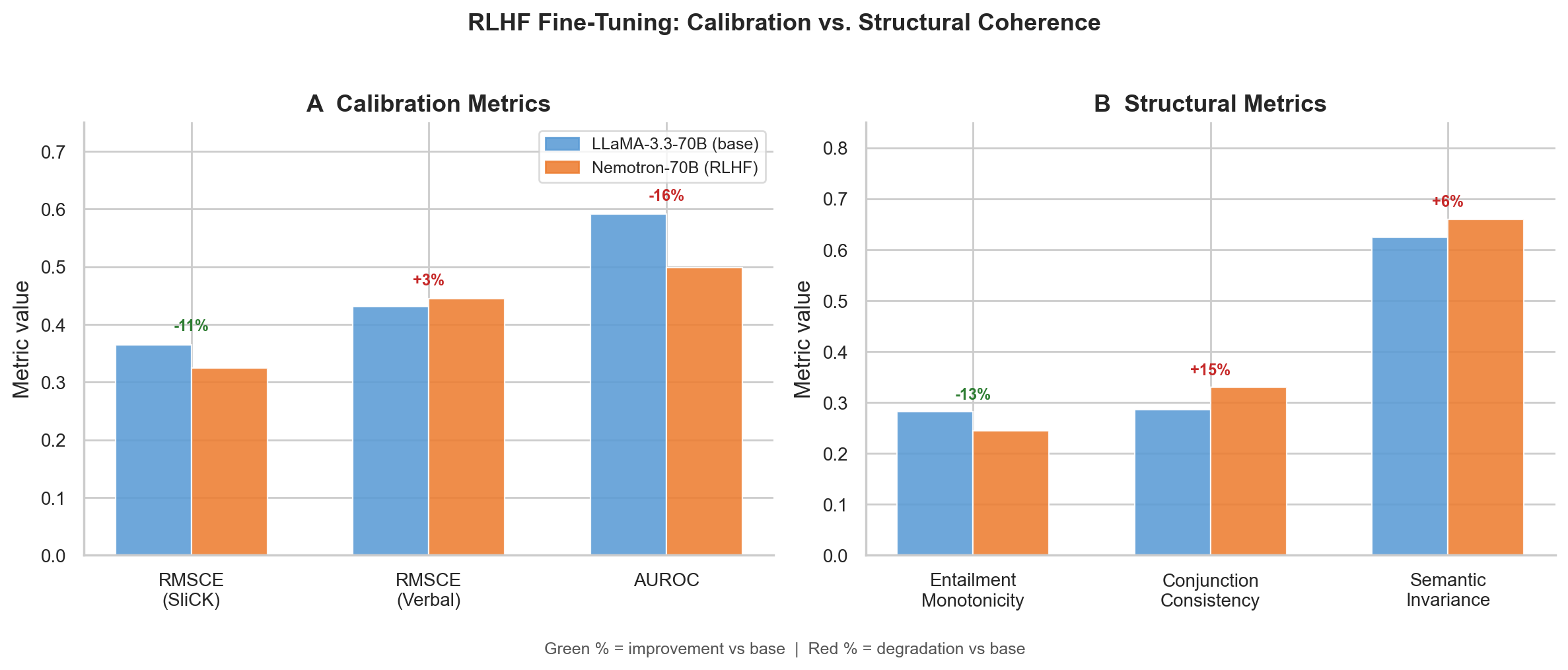}
    \caption{\textbf{Comparison against RLHF.} Impact of RLHF alignment across all metrics.}
    \label{fig:rlhf_comparison}
\end{figure}

\paragraph{RLHF.} Comparing Llama-3.1-70B against Nemotron-Llama-3.1-70B \citep{wang2024helpsteer2preference}, an RLHF-only variant, RLHF modestly improves SliCK calibration (RMSCE $0.365 \to 0.325$) but collapses discriminability to chance (AUROC $0.591 \to 0.498$) and worsens conjunction consistency by 15\%. Alignment optimizes for confident outputs at the direct expense of accurate uncertainty quantification and structural coherence.

\paragraph{Chain-of-Thought.} Adding the zero-shot CoT suffix (``Let's think step by step.'') to LLaMA-3-8B-Instruct improves RMSCE by 22\% ($0.271 \to 0.212$) and conjunction consistency by 41\% ($0.244 \to 0.144$), but leaves semantic invariance essentially unchanged ($0.498 \to 0.488$). Explicit reasoning reduces multi-hop overconfidence but it surprisingly cannot resolve prompt sensitivity.

\begin{figure}[t!]
\centering
\includegraphics[width=\columnwidth]{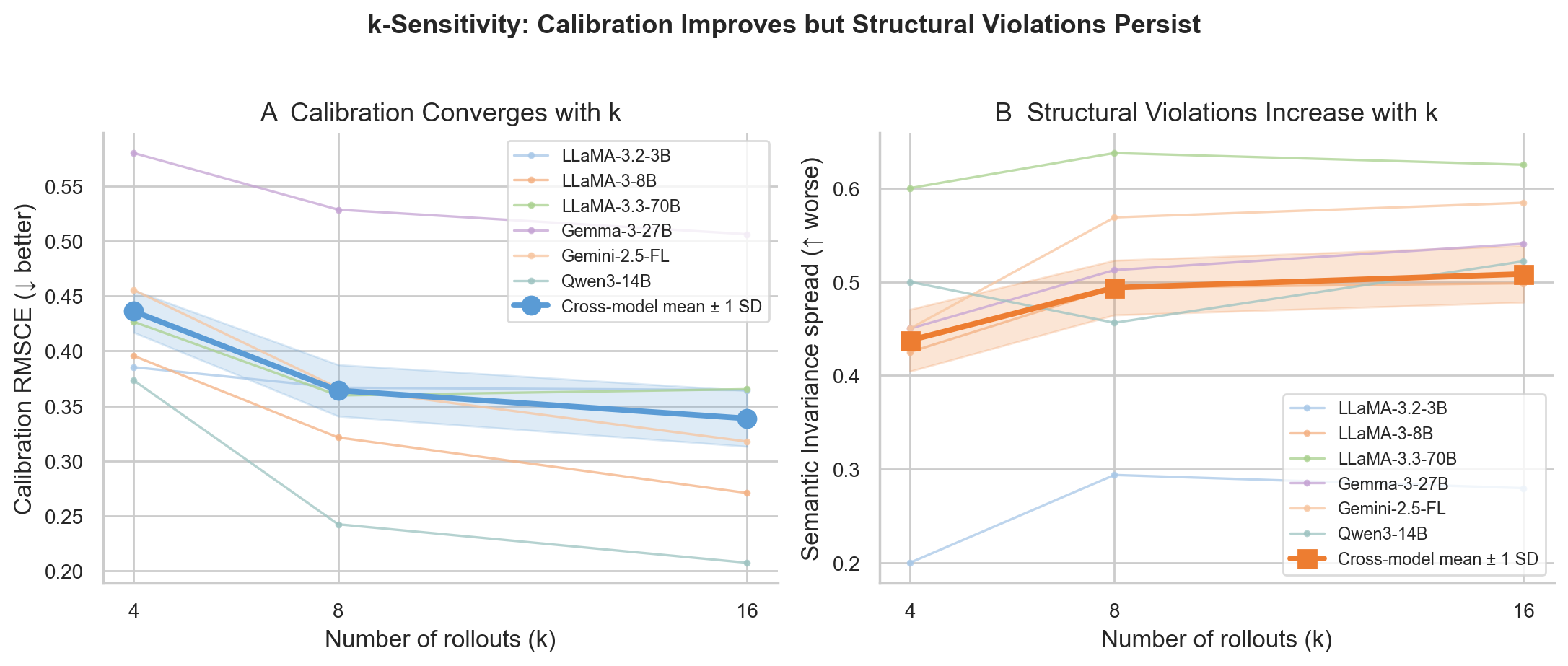}
\caption{\label{fig:ksen} As rollout count $k$ increases, calibration error (RMSCE) converges downward while semantic invariance violations increase, demonstrating that structural violations are a property of the model distribution rather than sampling noise.} 
\end{figure}

\paragraph{Sample Size (SliCK)} Subsampling our $k=16$ generations to $k \in \{4, 8, 16\}$ across six models, RMSCE decreases by 22\% ($0.436 \to 0.339$) while semantic invariance spread increases by 16\% ($0.438 \to 0.508$; Figure~\ref{fig:ksen}). More rollouts sharpen confidence estimates and expose structural violations rather than mitigate them, confirming the violations are properties of the model distribution, not sampling noise.

\section{Discussion}
\label{sec:conclusion}
 
We argued that calibration alone is poor for evaluating LLM confidence: it is satisfied by trivial constant predictors, depends on the evaluation distribution, and is silent about internal consistency. The \textbf{C1} metrics operationalize a richer evaluation along three axes: structural coherence, faithfulness, and usefulness.

%  \textbf{Why these properties?}
% Structural properties follow directly from probability axioms; violations imply confidence cannot be used consistently for decision-making. Faithfulness ensures invariance over semantic equivalence classes, and usefulness captures predictive performance. Our sampling-based protocol approximates the full distribution over answers, enabling evaluation of properties inaccessible to pointwise estimators.

% \textbf{Why it matters.}
% Confidence is used for abstention, routing, and aggregation, all of which assume probabilistic coherence. We show this assumption often fails: estimators can appear calibrated while violating basic structure, leading to unreliable decisions.

\textbf{Apparent calibration can reflect collapse.} Verbal and logit-based estimators concentrate scores near 1.0, which inflates apparent consistency on multiple structural and faithfulness properties. Single-number RMSCE cannot distinguish this regime from genuine coherence; SliCK's broader output range exposes structural violations the model itself harbors, including a 31\% entailment monotonicity violation rate and substantial conjunction inconsistency.
 
\textbf{Coherence and correctness do not scale together.} Within the LLaMA-3 family, prompt semantic invariance worsens with parameter count while RMSCE does not. Faithfulness and calibration are distinguishable axes, and a single metric cannot adjudicate between them.
 
\textbf{C1 as the new metrics for training.} Structural coherence may also be tractable as a training signal: fine-tuning directly against conjunction or entailment violations is unexplored. Whether RLHF degrades structural coherence in tandem with calibration \cite{openai2024gpt4technicalreport} is also an open question.

%% ============================================================
\section*{Impact Statement}
%% ============================================================

This paper presents work whose goal is to advance the field of Machine Learning. Improving the reliability of confidence estimation in LLMs has broad positive societal implications, enabling safer deployment of these systems in high-stakes applications. There are no specific ethical harms we feel must be highlighted here.

%% ============================================================
\bibliographystyle{icml2026} % Use the bst file provided by the workshop (often icml2026 or plainnat)
\bibliography{references} % Replace with your actual .bib file name

@misc{mazeika2025utilityengineeringanalyzingcontrolling,
      title={Utility Engineering: Analyzing and Controlling Emergent Value Systems in AIs}, 
      author={Mantas Mazeika and Xuwang Yin and Rishub Tamirisa and Jaehyuk Lim and Bruce W. Lee and Richard Ren and Long Phan and Norman Mu and Adam Khoja and Oliver Zhang and Dan Hendrycks},
      year={2025},
      eprint={2502.08640},
      archivePrefix={arXiv},
      primaryClass={cs.LG},
      url={https://arxiv.org/abs/2502.08640}, 
}

@misc{tian2023justaskcalibrationstrategies,
      title={Just Ask for Calibration: Strategies for Eliciting Calibrated Confidence Scores from Language Models Fine-Tuned with Human Feedback}, 
      author={Katherine Tian and Eric Mitchell and Allan Zhou and Archit Sharma and Rafael Rafailov and Huaxiu Yao and Chelsea Finn and Christopher D. Manning},
      year={2023},
      eprint={2305.14975},
      archivePrefix={arXiv},
      primaryClass={cs.CL},
      url={https://arxiv.org/abs/2305.14975}, 
}

@misc{kadavath2022languagemodelsmostlyknow,
      title={Language Models (Mostly) Know What They Know}, 
      author={Saurav Kadavath and Tom Conerly and Amanda Askell and Tom Henighan and Dawn Drain and Ethan Perez and Nicholas Schiefer and Zac Hatfield-Dodds and Nova DasSarma and Eli Tran-Johnson and Scott Johnston and Sheer El-Showk and Andy Jones and Nelson Elhage and Tristan Hume and Anna Chen and Yuntao Bai and Sam Bowman and Stanislav Fort and Deep Ganguli and Danny Hernandez and Josh Jacobson and Jackson Kernion and Shauna Kravec and Liane Lovitt and Kamal Ndousse and Catherine Olsson and Sam Ringer and Dario Amodei and Tom Brown and Jack Clark and Nicholas Joseph and Ben Mann and Sam McCandlish and Chris Olah and Jared Kaplan},
      year={2022},
      eprint={2207.05221},
      archivePrefix={arXiv},
      primaryClass={cs.CL},
      url={https://arxiv.org/abs/2207.05221}, 
}

@misc{gekhman2024doesfinetuningllmsnew,
      title={Does Fine-Tuning LLMs on New Knowledge Encourage Hallucinations?}, 
      author={Zorik Gekhman and Gal Yona and Roee Aharoni and Matan Eyal and Amir Feder and Roi Reichart and Jonathan Herzig},
      year={2024},
      eprint={2405.05904},
      archivePrefix={arXiv},
      primaryClass={cs.CL},
      url={https://arxiv.org/abs/2405.05904}, 
}

@misc{yang2025qwen3technicalreport,
      title={Qwen3 Technical Report}, 
      author={An Yang and Anfeng Li and Baosong Yang and Beichen Zhang and Binyuan Hui and Bo Zheng and Bowen Yu and Chang Gao and Chengen Huang and Chenxu Lv and Chujie Zheng and Dayiheng Liu and Fan Zhou and Fei Huang and Feng Hu and Hao Ge and Haoran Wei and Huan Lin and Jialong Tang and Jian Yang and Jianhong Tu and Jianwei Zhang and Jianxin Yang and Jiaxi Yang and Jing Zhou and Jingren Zhou and Junyang Lin and Kai Dang and Keqin Bao and Kexin Yang and Le Yu and Lianghao Deng and Mei Li and Mingfeng Xue and Mingze Li and Pei Zhang and Peng Wang and Qin Zhu and Rui Men and Ruize Gao and Shixuan Liu and Shuang Luo and Tianhao Li and Tianyi Tang and Wenbiao Yin and Xingzhang Ren and Xinyu Wang and Xinyu Zhang and Xuancheng Ren and Yang Fan and Yang Su and Yichang Zhang and Yinger Zhang and Yu Wan and Yuqiong Liu and Zekun Wang and Zeyu Cui and Zhenru Zhang and Zhipeng Zhou and Zihan Qiu},
      year={2025},
      eprint={2505.09388},
      archivePrefix={arXiv},
      primaryClass={cs.CL},
      url={https://arxiv.org/abs/2505.09388}, 
}

@misc{wei2024measuringshortformfactualitylarge,
      title={Measuring short-form factuality in large language models}, 
      author={Jason Wei and Nguyen Karina and Hyung Won Chung and Yunxin Joy Jiao and Spencer Papay and Amelia Glaese and John Schulman and William Fedus},
      year={2024},
      eprint={2411.04368},
      archivePrefix={arXiv},
      primaryClass={cs.CL},
      url={https://arxiv.org/abs/2411.04368}, 
}

@article{phan2025humanitysexam,
   title={A benchmark of expert-level academic questions to assess AI capabilities},
   volume={649},
   ISSN={1476-4687},
   url={http://dx.doi.org/10.1038/s41586-025-09962-4},
   DOI={10.1038/s41586-025-09962-4},
   number={8099},
   journal={Nature},
   publisher={Springer Science and Business Media LLC},
   author={Phan, Long and Gatti, Alice and Li, Nathaniel and Khoja, Adam and Kim, Ryan and Ren, Richard and Hausenloy, Jason and Zhang, Oliver and Mazeika, Mantas and Hendrycks, Dan and Han, Ziwen and Hu, Josephina and Zhang, Hugh and Zhang, Chen Bo Calvin and Shaaban, Mohamed and Ling, John and Shi, Sean and Choi, Michael and Agrawal, Anish and Chopra, Arnav and Nattanmai, Aakaash and McKellips, Gordon and Cheraku, Anish and Suhail, Asim and Luo, Ethan and Deng, Marvin and Luo, Jason and Zhang, Ashley and Jindel, Kavin and Paek, Jay and Halevy, Kasper and Baranov, Allen and Liu, Michael and Avadhanam, Advaith and Zhang, David and Cheng, Vincent and Ma, Brad and Fu, Evan and Do, Liam and Lass, Joshua and Yang, Hubert and Sunkari, Surya and Bharath, Vishruth and Ai, Violet and Leung, James and Agrawal, Rishit and Zhou, Alan and Chen, Kevin and Kalpathi, Tejas and Xu, Ziqi and Wang, Gavin and Xiao, Tyler and Maung, Erik and Lee, Sam and Yang, Ryan and Yue, Roy and Zhao, Ben and Yoon, Julia and Sun, Xiangwan and Singh, Aryan and Peng, Clark and Osbey, Tyler and Wang, Taozhi and Echeazu, Daryl and Wu, Timothy and Patel, Spandan and Kulkarni, Vidhi and Sundarapandiyan, Vijaykaarti and Le, Andrew and Nasim, Zafir and Yalam, Srikar and Kasamsetty, Ritesh and Samal, Soham and Sun, David and Shah, Nihar and Saha, Abhijeet and Zhang, Alex and Nguyen, Leon and Nagumalli, Laasya and Wang, Kaixin and Wu, Aidan and Telluri, Anwith and Yue, Summer and Wang, Alexandr and Dodonov, Dmitry and Nguyen, Tung and Lee, Jaeho and Anderson, Daron and Doroshenko, Mikhail and Stokes, Alun Cennyth and Mahmood, Mobeen and Pokutnyi, Oleksandr and Iskra, Oleg and Wang, Jessica P. and Levin, John-Clark and Kazakov, Mstyslav and Feng, Fiona and Feng, Steven Y. and Zhao, Haoran and Yu, Michael and Gangal, Varun and Zou, Chelsea and Wang, Zihan and Popov, Serguei and Gerbicz, Robert and Galgon, Geoff and Schmitt, Johannes and Yeadon, Will and Lee, Yongki and Sauers, Scott and Sanchez, Alvaro and Giska, Fabian and Roth, Marc and Riis, Søren and Utpala, Saiteja and Burns, Noah and Goshu, Gashaw M. and Naiya, Mohinder Maheshbhai and Agu, Chidozie and Giboney, Zachary and Cheatom, Antrell and Fournier-Facio, Francesco and Crowson, Sarah-Jane and Finke, Lennart and Cheng, Zerui and Zampese, Jennifer and Hoerr, Ryan G. and Nandor, Mark and Park, Hyunwoo and Gehrunger, Tim and Cai, Jiaqi and McCarty, Ben and Garretson, Alexis C. and Taylor, Edwin and Sileo, Damien and Ren, Qiuyu and Qazi, Usman and Li, Lianghui and Nam, Jungbae and Wydallis, John B. and Arkhipov, Pavel and Shi, Jack Wei Lun and Bacho, Aras and Willcocks, Chris G. and Cao, Hangrui and Motwani, Sumeet and de Oliveira Santos, Emily and Veith, Johannes and Vendrow, Edward and Cojoc, Doru and Zenitani, Kengo and Robinson, Joshua and Tang, Longke and Li, Yuqi and Vendrow, Joshua and Fraga, Natanael Wildner and Kuchkin, Vladyslav and Maksimov, Andrey Pupasov and Marion, Pierre and Efremov, Denis and Lynch, Jayson and Liang, Kaiqu and Mikov, Aleksandar and Gritsevskiy, Andrew and Guillod, Julien and Demir, Gözdenur and Martinez, Dakotah and Pageler, Ben and Zhou, Kevin and Soori, Saeed and Press, Ori and Tang, Henry and Rissone, Paolo and Green, Sean R. and Brüssel, Lina and Twayana, Moon and Dieuleveut, Aymeric and Imperial, Joseph Marvin and Prabhu, Ameya and Yang, Jinzhou and Crispino, Nick and Rao, Arun and Zvonkine, Dimitri and Loiseau, Gabriel and Kalinin, Mikhail and Lukas, Marco and Manolescu, Ciprian and Stambaugh, Nate and Mishra, Subrata and Hogg, Tad and Bosio, Carlo and Coppola, Brian P. and Salazar, Julian and Jin, Jaehyeok and Sayous, Rafael and Ivanov, Stefan and Schwaller, Philippe and Senthilkumar, Shaipranesh and Bran, Andres M. and Algaba, Andres and Van den Houte, Kelsey and Van Der Sypt, Lynn and Verbeken, Brecht and Noever, David and Kopylov, Alexei and Myklebust, Benjamin and Li, Bikun and Schut, Lisa and Zheltonozhskii, Evgenii and Yuan, Qiaochu and Lim, Derek and Stanley, Richard and Yang, Tong and Maar, John and Wykowski, Julian and Oller, Mart and Sahu, Anmol and Ardito, Cesare Giulio and Hu, Yuzheng and Kamdoum, Ariel Ghislain Kemogne and Jin, Alvin and Vilchis, Tobias Garcia and Zu, Yuexuan and Lackner, Martin and Koppel, James and Sun, Gongbo and Antonenko, Daniil S. and Chern, Steffi and Zhao, Bingchen and Arsene, Pierrot and Cavanagh, Joseph M. and Li, Daofeng and Shen, Jiawei and Crisostomi, Donato and Zhang, Wenjin and Dehghan, Ali and Ivanov, Sergey and Perrella, David and Kaparov, Nurdin and Zang, Allen and Sucholutsky, Ilia and Kharlamova, Arina and Orel, Daniil and Poritski, Vladislav and Ben-David, Shalev and Berger, Zachary and Whitfill, Parker and Foster, Michael and Munro, Daniel and Ho, Linh and Sivarajan, Shankar and Hava, Dan Bar and Kuchkin, Aleksey and Holmes, David and Rodriguez-Romero, Alexandra and Sommerhage, Frank and Zhang, Anji and Moat, Richard and Schneider, Keith and Kazibwe, Zakayo and Clarke, Don and Kim, Dae Hyun and Dias, Felipe Meneguitti and Fish, Sara and Elser, Veit and Kreiman, Tobias and Vilchis, Victor Efren Guadarrama and Klose, Immo and Anantheswaran, Ujjwala and Zweiger, Adam and Rawal, Kaivalya and Li, Jeffery and Nguyen, Jeremy and Daans, Nicolas and Heidinger, Haline and Radionov, Maksim and Rozhoň, Václav and Ginis, Vincent and Stump, Christian and Cohen, Niv and Poświata, Rafał and Tkadlec, Josef and Goldfarb, Alan and Wang, Chenguang and Padlewski, Piotr and Barzowski, Stanislaw and Montgomery, Kyle and Stendall, Ryan and Tucker-Foltz, Jamie and Stade, Jack and Rogers, T. Ryan and Goertzen, Tom and Grabb, Declan and Shukla, Abhishek and Givré, Alan and Ambay, John Arnold and Sen, Archan and Aziz, Muhammad Fayez and Inlow, Mark H. and He, Hao and Zhang, Ling and Kaddar, Younesse and Ängquist, Ivar and Chen, Yanxu and Wang, Harrison K. and Ramakrishnan, Kalyan and Thornley, Elliott and Terpin, Antonio and Schoelkopf, Hailey and Zheng, Eric and Carmi, Avishy and Brown, Ethan D. L. and Zhu, Kelin and Bartolo, Max and Wheeler, Richard and Stehberger, Martin and Bradshaw, Peter and Heimonen, JP and Sridhar, Kaustubh and Akov, Ido and Sandlin, Jennifer and Makarychev, Yury and Tam, Joanna and Hoang, Hieu and Cunningham, David M. and Goryachev, Vladimir and Patramanis, Demosthenes and Krause, Michael and Redenti, Andrew and Aldous, David and Lai, Jesyin and Coleman, Shannon and Xu, Jiangnan and Lee, Sangwon and Magoulas, Ilias and Zhao, Sandy and Tang, Ning and Cohen, Michael K. and Paradise, Orr and Kirchner, Jan Hendrik and Ovchynnikov, Maksym and Matos, Jason O. and Shenoy, Adithya and Wang, Michael and Nie, Yuzhou and Sztyber-Betley, Anna and Faraboschi, Paolo and Riblet, Robin and Crozier, Jonathan and Halasyamani, Shiv and Verma, Shreyas and Joshi, Prashant and Meril, Eli and Ma, Ziqiao and Andréoletti, Jérémy and Singhal, Raghav and Platnick, Jacob and Nevirkovets, Volodymyr and Basler, Luke and Ivanov, Alexander and Khoury, Seri and Gustafsson, Nils and Piccardo, Marco and Mostaghimi, Hamid and Chen, Qijia and Singh, Virendra and Khánh, Tran Quoc and Rosu, Paul and Szlyk, Hannah and Brown, Zachary and Narayan, Himanshu and Menezes, Aline and Roberts, Jonathan and Alley, William and Sun, Kunyang and Patel, Arkil and Lamparth, Max and Reuel, Anka and Xin, Linwei and Xu, Hanmeng and Loader, Jacob and Martin, Freddie and Wang, Zixuan and Achilleos, Andrea and Preu, Thomas and Korbak, Tomek and Bosio, Ida and Kazemi, Fereshteh and Chen, Ziye and Bálint, Biró and Lo, Eve J. Y. and Wang, Jiaqi and Nunes, Maria Inês S. and Milbauer, Jeremiah and Bari, M. Saiful and Wang, Zihao and Ansarinejad, Behzad and Sun, Yewen and Durand, Stephane and Elgnainy, Hossam and Douville, Guillaume and Tordera, Daniel and Balabanian, George and Wolff, Hew and Kvistad, Lynna and Milliron, Hsiaoyun and Sakor, Ahmad and Eron, Murat and Favre, Andrew and Shah, Shailesh and Zhou, Xiaoxiang and Kamalov, Firuz and Abdoli, Sherwin and Santens, Tim and Barkan, Shaul and Tee, Allison and Zhang, Robin and Tomasiello, Alessandro and De Luca, G. Bruno and Looi, Shi-Zhuo and Le, Vinh-Kha and Kolt, Noam and Pan, Jiayi and Rodman, Emma and Drori, Jacob and Fossum, Carl J. and Muennighoff, Niklas and Jagota, Milind and Pradeep, Ronak and Fan, Honglu and Eicher, Jonathan and Chen, Michael and Thaman, Kushal and Merrill, William and Firsching, Moritz and Harris, Carter and Ciobâcă, Stefan and Gross, Jason and Pandey, Rohan and Gusev, Ilya and Jones, Adam and Agnihotri, Shashank and Zhelnov, Pavel and Mofayezi, Mohammadreza and Piperski, Alexander and Zhang, David K. and Dobarskyi, Kostiantyn and Leventov, Roman and Soroko, Ignat and Duersch, Joshua and Taamazyan, Vage and Ho, Andrew and Ma, Wenjie and Held, William and Xian, Ruicheng and Zebaze, Armel Randy and Mohamed, Mohanad and Leser, Julian Noah and Yuan, Michelle X. and Yacar, Laila and Lengler, Johannes and Olszewska, Katarzyna and Di Fratta, Claudio and Oliveira, Edson and Jackson, Joseph W. and Zou, Andy and Chidambaram, Muthu and Manik, Timothy and Haffenden, Hector and Stander, Dashiell and Dasouqi, Ali and Shen, Alexander and Golshani, Bita and Stap, David and Kretov, Egor and Uzhou, Mikalai and Zhidkovskaya, Alina Borisovna and Winter, Nick and Rodriguez, Miguel Orbegozo and Lauff, Robert and Wehr, Dustin and Tang, Colin and Hossain, Zaki and Phillips, Shaun and Samuele, Fortuna and Ekström, Fredrik and Hammon, Angela and Patel, Oam and Farhidi, Faraz and Medley, George and Mohammadzadeh, Forough and Peñaflor, Madellene and Kassahun, Haile and Friedrich, Alena and Perez, Rayner Hernandez and Pyda, Daniel and Sakal, Taom and Dhamane, Omkar and Mirabadi, Ali Khajegili and Hallman, Eric and Okutsu, Kenchi and Battaglia, Mike and Maghsoudimehrabani, Mohammad and Amit, Alon and Hulbert, Dave and Pereira, Roberto and Weber, Simon and Handoko and Peristyy, Anton and Malina, Stephen and Mehkary, Mustafa and Aly, Rami and Reidegeld, Frank and Dick, Anna-Katharina and Friday, Cary and Singh, Mukhwinder and Shapourian, Hassan and Kim, Wanyoung and Costa, Mariana and Gurdogan, Hubeyb and Kumar, Harsh and Ceconello, Chiara and Zhuang, Chao and Park, Haon and Carroll, Micah and Tawfeek, Andrew R. and Steinerberger, Stefan and Aggarwal, Daattavya and Kirchhof, Michael and Dai, Linjie and Kim, Evan and Ferret, Johan and Shah, Jainam and Wang, Yuzhou and Yan, Minghao and Burdzy, Krzysztof and Zhang, Lixin and Franca, Antonio and Pham, Diana T. and Loh, Kang Yong and Robinson, Joshua and Jackson, Abram and Giordano, Paolo and Petersen, Philipp and Cosma, Adrian and Colino, Jesus and White, Colin and Votava, Jacob and Vinnikov, Vladimir and Delaney, Ethan and Spelda, Petr and Stritecky, Vit and Shahid, Syed M. and Mourrat, Jean-Christophe and Vetoshkin, Lavr and Sponselee, Koen and Bacho, Renas and Yong, Zheng-Xin and de la Rosa, Florencia and Cho, Nathan and Li, Xiuyu and Malod, Guillaume and Weller, Orion and Albani, Guglielmo and Lang, Leon and Laurendeau, Julien and Kazakov, Dmitry and Adesanya, Fatimah and Portier, Julien and Hollom, Lawrence and Souza, Victor and Zhou, Yuchen Anna and Degorre, Julien and Yaln, Yiğit and Obikoya, Gbenga Daniel and Michael Pokorny, Rai and Bigi, Filippo and Boscá, M. C. and Shumar, Oleg and Bacho, Kaniuar and Recchia, Gabriel and Popescu, Mara and Shulga, Nikita and Tanwie, Ngefor Mildred and Lux, Thomas C. H. and Rank, Ben and Ni, Colin and Brooks, Matthew and Yakimchyk, Alesia and Quinn Liu, Huanxu and Cavalleri, Stefano and Häggström, Olle and Verkama, Emil and Newbould, Joshua and Gundlach, Hans and Brito-Santana, Leonor and Amaro, Brian and Vajipey, Vivek and Grover, Rynaa and Wang, Ting and Kratish, Yosi and Li, Wen-Ding and Gopi, Sivakanth and Caciolai, Andrea and de Witt, Christian Schroeder and Hernández-Cámara, Pablo and Rodolà, Emanuele and Robins, Jules and Williamson, Dominic and Raynor, Brad and Qi, Hao and Segev, Ben and Fan, Jingxuan and Martinson, Sarah and Wang, Erik Y. and Hausknecht, Kaylie and Brenner, Michael P. and Mao, Mao and Demian, Christoph and Kassani, Peyman and Zhang, Xinyu and Avagian, David and Scipio, Eshawn Jessica and Ragoler, Alon and Tan, Justin and Sims, Blake and Plecnik, Rebeka and Kirtland, Aaron and Bodur, Omer Faruk and Shinde, D. P. and Labrador, Yan Carlos Leyva and Adoul, Zahra and Zekry, Mohamed and Karakoc, Ali and Santos, Tania C. B. and Shamseldeen, Samir and Karim, Loukmane and Liakhovitskaia, Anna and Resman, Nate and Farina, Nicholas and Gonzalez, Juan Carlos and Maayan, Gabe and Anderson, Earth and De Oliveira Pena, Rodrigo and Kelley, Elizabeth and Mariji, Hodjat and Pouriamanesh, Rasoul and Wu, Wentao and Finocchio, Ross and Alarab, Ismail and Cole, Joshua and Ferreira, Danyelle and Johnson, Bryan and Safdari, Mohammad and Dai, Liangti and Arthornthurasuk, Siriphan and McAlister, Isaac C. and Moyano, Alejandro José and Pronin, Alexey and Fan, Jing and Ramirez-Trinidad, Angel and Malysheva, Yana and Pottmaier, Daphiny and Taheri, Omid and Stepanic, Stanley and Perry, Samuel and Askew, Luke and Rodrguez, Raúl Adrián Huerta and Minissi, Ali M. R. and Lorena, Ricardo and Iyer, Krishnamurthy and Fasiludeen, Arshad Anil and Clark, Ronald and Ducey, Josh and Piza, Matheus and Somrak, Maja and Vergo, Eric and Qin, Juehang and Borbás, Benjámin and Chu, Eric and Lindsey, Jack and Jallon, Antoine and McInnis, I. M. J. and Chen, Evan and Semler, Avi and Gloor, Luk and Shah, Tej and Carauleanu, Marc and Lauer, Pascal and Huy, Tran Duc and Shahrtash, Hossein and Duc, Emilien and Lewark, Lukas and Brown, Assaf and Albanie, Samuel and Weber, Brian and Vaz, Warren S. and Clavier, Pierre and Fan, Yiyang and Poesia Reis e Silva, Gabriel and Tony Lian, Long and Abramovitch, Marcus and Jiang, Xi and Mendoza, Sandra and Islam, Murat and Gonzalez, Juan and Mavroudis, Vasilios and Xu, Justin and Kumar, Pawan and Goswami, Laxman Prasad and Bugas, Daniel and Heydari, Nasser and Jeanplong, Ferenc and Jansen, Thorben and Pinto, Antonella and Apronti, Archimedes and Galal, Abdallah and Ze-An, Ng and Singh, Ankit and Jiang, Tong and of Arc Xavier, Joan and Agarwal, Kanu Priya and Berkani, Mohammed and Zhang, Gang and Du, Zhehang and de Oliveira Junior, Benedito Alves and Malishev, Dmitry and Remy, Nicolas and Hartman, Taylor D. and Tarver, Tim and Mensah, Stephen and Loume, Gautier Abou and Morak, Wiktor and Habibi, Farzad and Hoback, Sarah and Cai, Will and Gimenez, Javier and Montecillo, Roselynn Grace and Łucki, Jakub and Campbell, Russell and Sharma, Asankhaya and Meer, Khalida and Gul, Shreen and Gonzalez, Daniel Espinosa and Alapont, Xavier and Hoover, Alex and Chhablani, Gunjan and Vargus, Freddie and Agarwal, Arunim and Jiang, Yibo and Patil, Deepakkumar and Outevsky, David and Scaria, Kevin Joseph and Maheshwari, Rajat and Dendane, Abdelkader and Shukla, Priti and Cartwright, Ashley and Bogdanov, Sergei and Mündler, Niels and Möller, Sören and Arnaboldi, Luca and Thaman, Kunvar and Siddiqi, Muhammad Rehan and Saxena, Prajvi and Gupta, Himanshu and Fruhauff, Tony and Sherman, Glen and Vincze, Mátyás and Usawasutsakorn, Siranut and Ler, Dylan and Radhakrishnan, Anil and Enyekwe, Innocent and Salauddin, Sk Md and Muzhen, Jiang and Maksapetyan, Aleksandr and Rossbach, Vivien and Harjadi, Chris and Bahaloohoreh, Mohsen and Sparrow, Claire and Sidhu, Jasdeep and Ali, Sam and Bian, Song and Lai, John and Singer, Eric and Uro, Justine Leon and Bateman, Greg and Sayed, Mohamed and Menshawy, Ahmed and Duclosel, Darling and Bezzi, Dario and Jain, Yashaswini and Aaron, Ashley and Tiryakioglu, Murat and Siddh, Sheeshram and Krenek, Keith and Shah, Imad Ali and Jin, Jun and Creighton, Scott and Peskoff, Denis and EL-Wasif, Zienab and P, Ragavendran and Richmond, Michael and McGowan, Joseph and Patwardhan, Tejal and Sun, Hao-Yu and Sun, Ting and Zubić, Nikola and Sala, Samuele and Ebert, Stephen and Kaddour, Jean and Schottdorf, Manuel and Wang, Dianzhuo and Petruzella, Gerol and Meiburg, Alex and Medved, Tilen and ElSheikh, Ali and Hebbar, S. Ashwin and Vaquero, Lorenzo and Yang, Xianjun and Poulos, Jason and Zouhar, Vilém and Bogdanik, Sergey and Zhang, Mingfang and Sanz-Ros, Jorge and Anugraha, David and Dai, Yinwei and Nhu, Anh N. and Wang, Xue and Demircali, Ali Anil and Jia, Zhibai and Zhou, Yuyin and Wu, Juncheng and He, Mike and Chandok, Nitin and Sinha, Aarush and Luo, Gaoxiang and Le, Long and Noyé, Mickaël and Perełkiewicz, Michał and Pantidis, Ioannis and Qi, Tianbo and Purohit, Soham Sachin and Parcalabescu, Letitia and Nguyen, Thai-Hoa and Winata, Genta Indra and Ponti, Edoardo M. and Li, Hanchen and Dhole, Kaustubh and Park, Jongee and Abbondanza, Dario and Wang, Yuanli and Nayak, Anupam and Caetano, Diogo M. and Wong, Antonio A. W. L. and del Rio-Chanona, Maria and Kondor, Dániel and Francois, Pieter and Chalstrey, Ed and Zsambok, Jakob and Hoyer, Dan and Reddish, Jenny and Hauser, Jakob and Rodrigo-Ginés, Francisco-Javier and Datta, Suchandra and Shepherd, Maxwell and Kamphuis, Thom and Zhang, Qizheng and Kim, Hyunjun and Sun, Ruiji and Yao, Jianzhu and Dernoncourt, Franck and Krishna, Satyapriya and Rismanchian, Sina and Pu, Bonan and Pinto, Francesco and Wang, Yingheng and Shridhar, Kumar and Overholt, Kalon J. and Briia, Glib and Nguyen, Hieu and Quod Soler Bartomeu, David and Pang, Tony CY and Wecker, Adam and Xiong, Yifan and Li, Fanfei and Huber, Lukas S. and Jaeger, Joshua and De Maddalena, Romano and Lù, Xing Han and Zhang, Yuhui and Beger, Claas and Kon, Patrick Tser Jern and Li, Sean and Sanker, Vivek and Yin, Ming and Liang, Yihao and Zhang, Xinlu and Agrawal, Ankit and Yifei, Li S. and Zhang, Zechen and Cai, Mu and Sonmez, Yasin and Cozianu, Costin and Li, Changhao and Slen, Alex and Yu, Shoubin and Park, Hyun Kyu and Sarti, Gabriele and Briański, Marcin and Stolfo, Alessandro and Nguyen, Truong An and Zhang, Mike and Perlitz, Yotam and Hernandez-Orallo, Jose and Li, Runjia and Shabani, Amin and Juefei-Xu, Felix and Dhingra, Shikhar and Zohar, Orr and Nguyen, My Chiffon and Pondaven, Alexander and Yilmaz, Abdurrahim and Zhao, Xuandong and Jin, Chuanyang and Jiang, Muyan and Todoran, Stefan and Han, Xinyao and Kreuer, Jules and Rabern, Brian and Plassart, Anna and Maggetti, Martino and Yap, Luther and Geirhos, Robert and Kean, Jonathon and Wang, Dingsu and Mollaei, Sina and Sun, Chenkai and Yin, Yifan and Wang, Shiqi and Li, Rui and Chang, Yaowen and Wei, Anjiang and Bizeul, Alice and Wang, Xiaohan and Arrais, Alexandre Oliveira and Mukherjee, Kushin and Chamorro-Padial, Jorge and Liu, Jiachen and Qu, Xingyu and Guan, Junyi and Bouyamourn, Adam and Wu, Shuyu and Plomecka, Martyna and Chen, Junda and Tang, Mengze and Deng, Jiaqi and Subramanian, Shreyas and Xi, Haocheng and Chen, Haoxuan and Zhang, Weizhi and Ren, Yinuo and Tu, Haoqin and Kim, Sejong and Chen, Yushun and Marjanović, Sara Vera and Ha, Junwoo and Luczyna, Grzegorz and Ma, Jeff J. and Shen, Zewen and Song, Dawn and Zhang, Cedegao E. and Wang, Zhun and Gendron, Gaël and Xiao, Yunze and Smucker, Leo and Weng, Erica and Lee, Kwok Hao and Ye, Zhe and Ermon, Stefano and Lopez-Miguel, Ignacio D. and Knights, Theo and Gitter, Anthony and Park, Namkyu and Wei, Boyi and Chen, Hongzheng and Pai, Kunal and Elkhanany, Ahmed and Lin, Han and Siedler, Philipp D. and Fang, Jichao and Mishra, Ritwik and Zsolnai-Fehér, Károly and Jiang, Xilin and Khan, Shadab and Yuan, Jun and Jain, Rishab Kumar and Lin, Xi and Peterson, Mike and Wang, Zhe and Malusare, Aditya and Tang, Maosen and Gupta, Isha and Fosin, Ivan and Kang, Timothy and Dworakowska, Barbara and Matsumoto, Kazuki and Zheng, Guangyao and Sewuster, Gerben and Villanueva, Jorge Pretel and Rannev, Ivan and Chernyavsky, Igor and Chen, Jiale and Banik, Deepayan and Racz, Ben and Dong, Wenchao and Wang, Jianxin and Bashmal, Laila and Gonçalves, Duarte V. and Hu, Wei and Bar, Kaushik and Bohdal, Ondrej and Patlan, Atharv Singh and Dhuliawala, Shehzaad and Geirhos, Caroline and Wist, Julien and Kansal, Yuval and Chen, Bingsen and Tire, Kutay and Yücel, Atak Talay and Christof, Brandon and Singla, Veerupaksh and Song, Zijian and Chen, Sanxing and Ge, Jiaxin and Ponkshe, Kaustubh and Park, Isaac and Shi, Tianneng and Ma, Martin Q. and Mak, Joshua and Lai, Sherwin and Moulin, Antoine and Cheng, Zhuo and Zhu, Zhanda and Zhang, Ziyi and Patil, Vaidehi and Jha, Ketan and Men, Qiutong and Wu, Jiaxuan and Zhang, Tianchi and Vieira, Bruno Hebling and Aji, Alham Fikri and Chung, Jae-Won and Mahfoud, Mohammed and Thi Hoang, Ha and Sperzel, Marc and Hao, Wei and Meding, Kristof and Xu, Sihan and Kostakos, Vassilis and Manini, Davide and Liu, Yueying and Toukmaji, Christopher and Yu, Eunmi and Demircali, Arif Engin and Sun, Zhiyi and Dewerpe, Ivan and Qin, Hongsen and Pflugfelder, Roman and Bailey, James and Morris, Johnathan and Heilala, Ville and Rosset, Sybille and Yu, Zishun and Chen, Peter E. and Yeo, Woongyeong and Jain, Eeshaan and Chigurupati, Sreekar and Chernyavsky, Julia and Reddy, Sai Prajwal and Venugopalan, Subhashini and Batra, Hunar and Park, Core Francisco and Tran, Hieu and Maximiano, Guilherme and Zhang, Genghan and Liang, Yizhuo and Shiyu, Hu and Xu, Rongwu and Pan, Rui and Suresh, Siddharth and Liu, Ziqi and Gulati, Samaksh and Zhang, Songyang and Turchin, Peter and Bartlett, Christopher W. and Scotese, Christopher R. and Cao, Phuong M. and Wu, Ben and Karwowski, Jacek and Scaramuzza, Davide},
   year={2026},
   month=jan, pages={1139–1146} }

@article{trivedi2022musaboringmultihop,
  author       = {Harsh Trivedi and
                  Niranjan Balasubramanian and
                  Tushar Khot and
                  Ashish Sabharwal},
  title        = {MuSiQue: Multi-hop Questions via Single-hop Question Composition},
  journal      = {CoRR},
  volume       = {abs/2108.00573},
  year         = {2021},
  url          = {https://arxiv.org/abs/2108.00573},
  eprinttype   = {arXiv},
  eprint       = {2108.00573},
  bibsource    = {dblp computer science bibliography, https://dblp.org}
}

@misc{elazar2021measuringimprovingconsistencypretrained,
      title={Measuring and Improving Consistency in Pretrained Language Models}, 
      author={Yanai Elazar and Nora Kassner and Shauli Ravfogel and Abhilasha Ravichander and Eduard Hovy and Hinrich Schütze and Yoav Goldberg},
      year={2021},
      eprint={2102.01017},
      archivePrefix={arXiv},
      primaryClass={cs.CL},
      url={https://arxiv.org/abs/2102.01017}, 
}

@misc{geng2024surveyconfidenceestimationcalibration,
      title={A Survey of Confidence Estimation and Calibration in Large Language Models}, 
      author={Jiahui Geng and Fengyu Cai and Yuxia Wang and Heinz Koeppl and Preslav Nakov and Iryna Gurevych},
      year={2024},
      eprint={2311.08298},
      archivePrefix={arXiv},
      primaryClass={cs.CL},
      url={https://arxiv.org/abs/2311.08298}, 
}

@misc{openai2024gpt4technicalreport,
      title={GPT-4 Technical Report}, 
      author={OpenAI and Josh Achiam and Steven Adler and Sandhini Agarwal and Lama Ahmad and Ilge Akkaya and Florencia Leoni Aleman and Diogo Almeida and Janko Altenschmidt and Sam Altman and Shyamal Anadkat and Red Avila and Igor Babuschkin and Suchir Balaji and Valerie Balcom and Paul Baltescu and Haiming Bao and Mohammad Bavarian and Jeff Belgum and Irwan Bello and Jake Berdine and Gabriel Bernadett-Shapiro and Christopher Berner and Lenny Bogdonoff and Oleg Boiko and Madelaine Boyd and Anna-Luisa Brakman and Greg Brockman and Tim Brooks and Miles Brundage and Kevin Button and Trevor Cai and Rosie Campbell and Andrew Cann and Brittany Carey and Chelsea Carlson and Rory Carmichael and Brooke Chan and Che Chang and Fotis Chantzis and Derek Chen and Sully Chen and Ruby Chen and Jason Chen and Mark Chen and Ben Chess and Chester Cho and Casey Chu and Hyung Won Chung and Dave Cummings and Jeremiah Currier and Yunxing Dai and Cory Decareaux and Thomas Degry and Noah Deutsch and Damien Deville and Arka Dhar and David Dohan and Steve Dowling and Sheila Dunning and Adrien Ecoffet and Atty Eleti and Tyna Eloundou and David Farhi and Liam Fedus and Niko Felix and Simón Posada Fishman and Juston Forte and Isabella Fulford and Leo Gao and Elie Georges and Christian Gibson and Vik Goel and Tarun Gogineni and Gabriel Goh and Rapha Gontijo-Lopes and Jonathan Gordon and Morgan Grafstein and Scott Gray and Ryan Greene and Joshua Gross and Shixiang Shane Gu and Yufei Guo and Chris Hallacy and Jesse Han and Jeff Harris and Yuchen He and Mike Heaton and Johannes Heidecke and Chris Hesse and Alan Hickey and Wade Hickey and Peter Hoeschele and Brandon Houghton and Kenny Hsu and Shengli Hu and Xin Hu and Joost Huizinga and Shantanu Jain and Shawn Jain and Joanne Jang and Angela Jiang and Roger Jiang and Haozhun Jin and Denny Jin and Shino Jomoto and Billie Jonn and Heewoo Jun and Tomer Kaftan and Łukasz Kaiser and Ali Kamali and Ingmar Kanitscheider and Nitish Shirish Keskar and Tabarak Khan and Logan Kilpatrick and Jong Wook Kim and Christina Kim and Yongjik Kim and Jan Hendrik Kirchner and Jamie Kiros and Matt Knight and Daniel Kokotajlo and Łukasz Kondraciuk and Andrew Kondrich and Aris Konstantinidis and Kyle Kosic and Gretchen Krueger and Vishal Kuo and Michael Lampe and Ikai Lan and Teddy Lee and Jan Leike and Jade Leung and Daniel Levy and Chak Ming Li and Rachel Lim and Molly Lin and Stephanie Lin and Mateusz Litwin and Theresa Lopez and Ryan Lowe and Patricia Lue and Anna Makanju and Kim Malfacini and Sam Manning and Todor Markov and Yaniv Markovski and Bianca Martin and Katie Mayer and Andrew Mayne and Bob McGrew and Scott Mayer McKinney and Christine McLeavey and Paul McMillan and Jake McNeil and David Medina and Aalok Mehta and Jacob Menick and Luke Metz and Andrey Mishchenko and Pamela Mishkin and Vinnie Monaco and Evan Morikawa and Daniel Mossing and Tong Mu and Mira Murati and Oleg Murk and David Mély and Ashvin Nair and Reiichiro Nakano and Rajeev Nayak and Arvind Neelakantan and Richard Ngo and Hyeonwoo Noh and Long Ouyang and Cullen O'Keefe and Jakub Pachocki and Alex Paino and Joe Palermo and Ashley Pantuliano and Giambattista Parascandolo and Joel Parish and Emy Parparita and Alex Passos and Mikhail Pavlov and Andrew Peng and Adam Perelman and Filipe de Avila Belbute Peres and Michael Petrov and Henrique Ponde de Oliveira Pinto and Michael and Pokorny and Michelle Pokrass and Vitchyr H. Pong and Tolly Powell and Alethea Power and Boris Power and Elizabeth Proehl and Raul Puri and Alec Radford and Jack Rae and Aditya Ramesh and Cameron Raymond and Francis Real and Kendra Rimbach and Carl Ross and Bob Rotsted and Henri Roussez and Nick Ryder and Mario Saltarelli and Ted Sanders and Shibani Santurkar and Girish Sastry and Heather Schmidt and David Schnurr and John Schulman and Daniel Selsam and Kyla Sheppard and Toki Sherbakov and Jessica Shieh and Sarah Shoker and Pranav Shyam and Szymon Sidor and Eric Sigler and Maddie Simens and Jordan Sitkin and Katarina Slama and Ian Sohl and Benjamin Sokolowsky and Yang Song and Natalie Staudacher and Felipe Petroski Such and Natalie Summers and Ilya Sutskever and Jie Tang and Nikolas Tezak and Madeleine B. Thompson and Phil Tillet and Amin Tootoonchian and Elizabeth Tseng and Preston Tuggle and Nick Turley and Jerry Tworek and Juan Felipe Cerón Uribe and Andrea Vallone and Arun Vijayvergiya and Chelsea Voss and Carroll Wainwright and Justin Jay Wang and Alvin Wang and Ben Wang and Jonathan Ward and Jason Wei and CJ Weinmann and Akila Welihinda and Peter Welinder and Jiayi Weng and Lilian Weng and Matt Wiethoff and Dave Willner and Clemens Winter and Samuel Wolrich and Hannah Wong and Lauren Workman and Sherwin Wu and Jeff Wu and Michael Wu and Kai Xiao and Tao Xu and Sarah Yoo and Kevin Yu and Qiming Yuan and Wojciech Zaremba and Rowan Zellers and Chong Zhang and Marvin Zhang and Shengjia Zhao and Tianhao Zheng and Juntang Zhuang and William Zhuk and Barret Zoph},
      year={2024},
      eprint={2303.08774},
      archivePrefix={arXiv},
      primaryClass={cs.CL},
      url={https://arxiv.org/abs/2303.08774}, 
}

@INCOLLECTION{ramsey1926,
title = {Truth and Probability},
author = {Ramsey, Frank P.},
year = {1926},
chapter = {7},
pages = {156-198},
booktitle = {The Foundations of Mathematics and other Logical Essays},
editor = {Braithwaite, R. B.},
publisher = {McMaster University Archive for the History of Economic Thought},
url = {https://EconPapers.repec.org/RePEc:hay:hetcha:ramsey1926}
}

@article{Cox1946-COXPFA-3,
	author = {Richard T. Cox},
	doi = {10.2307/2272983},
	journal = {Journal of Symbolic Logic},
	number = {2},
	pages = {398--399},
	title = {Probability, Frequency and Reasonable Expectation},
	volume = {37},
	year = {1946}
}

@misc{zhou2024surveylargelanguagemodels,
      title={A Survey of Large Language Models in Medicine: Progress, Application, and Challenge}, 
      author={Hongjian Zhou and Fenglin Liu and Boyang Gu and Xinyu Zou and Jinfa Huang and Jinge Wu and Yiru Li and Sam S. Chen and Peilin Zhou and Junling Liu and Yining Hua and Chengfeng Mao and Chenyu You and Xian Wu and Yefeng Zheng and Lei Clifton and Zheng Li and Jiebo Luo and David A. Clifton},
      year={2024},
      eprint={2311.05112},
      archivePrefix={arXiv},
      primaryClass={cs.CL},
      url={https://arxiv.org/abs/2311.05112}, 
}

@misc{ye2024assessingcreativityllmsproposing,
      title={Assessing the Creativity of LLMs in Proposing Novel Solutions to Mathematical Problems}, 
      author={Junyi Ye and Jingyi Gu and Xinyun Zhao and Wenpeng Yin and Guiling Wang},
      year={2024},
      eprint={2410.18336},
      archivePrefix={arXiv},
      primaryClass={cs.CL},
      url={https://arxiv.org/abs/2410.18336}, 
}

@misc{lin2022teachingmodelsexpressuncertainty,
      title={Teaching Models to Express Their Uncertainty in Words}, 
      author={Stephanie Lin and Jacob Hilton and Owain Evans},
      year={2022},
      eprint={2205.14334},
      archivePrefix={arXiv},
      primaryClass={cs.CL},
      url={https://arxiv.org/abs/2205.14334}, 
}

@misc{xiong2024llmsexpressuncertaintyempirical,
      title={Can LLMs Express Their Uncertainty? An Empirical Evaluation of Confidence Elicitation in LLMs}, 
      author={Miao Xiong and Zhiyuan Hu and Xinyang Lu and Yifei Li and Jie Fu and Junxian He and Bryan Hooi},
      year={2024},
      eprint={2306.13063},
      archivePrefix={arXiv},
      primaryClass={cs.CL},
      url={https://arxiv.org/abs/2306.13063}, 
}

@misc{wang2024helpsteer2preference,
      title={HelpSteer2-Preference: Complementing Ratings with Preferences}, 
      author={Zhilin Wang and Yi Dong and Olivier Delalleau and Jiaqi Zeng and Gerald Shen and Daniel Egert and Jimmy J. Lin and Wei Ping and Yi Sun and Ming-Wei Chang},
      year={2024},
      eprint={2410.01257},
      archivePrefix={arXiv},
      primaryClass={cs.LG},
      url={https://arxiv.org/abs/2410.01257}
}

@article{lamb2026tokentemp,
  author    = {Lamb, Tom A. and Ivanova, Desi R. and Torr, Philip H. S. and Rudner, Tim G. J.},
  title     = {Improving Semantic Uncertainty Quantification in Language Model
               Question-Answering via Token-Level Temperature Scaling},
  journal   = {arXiv preprint arXiv:2604.07172},
  year      = {2026},
  url       = {https://arxiv.org/abs/2604.07172},
}

@misc{wang2023selfconsistencyimproveschainthought,
      title={Self-Consistency Improves Chain of Thought Reasoning in Language Models}, 
      author={Xuezhi Wang and Jason Wei and Dale Schuurmans and Quoc Le and Ed Chi and Sharan Narang and Aakanksha Chowdhery and Denny Zhou},
      year={2023},
      eprint={2203.11171},
      archivePrefix={arXiv},
      primaryClass={cs.CL},
      url={https://arxiv.org/abs/2203.11171}, 
}

@misc{manakul2023selfcheckgptzeroresourceblackboxhallucination,
      title={SelfCheckGPT: Zero-Resource Black-Box Hallucination Detection for Generative Large Language Models}, 
      author={Potsawee Manakul and Adian Liusie and Mark J. F. Gales},
      year={2023},
      eprint={2303.08896},
      archivePrefix={arXiv},
      primaryClass={cs.CL},
      url={https://arxiv.org/abs/2303.08896}, 
}

@misc{kuhn2023semanticuncertaintylinguisticinvariances,
      title={Semantic Uncertainty: Linguistic Invariances for Uncertainty Estimation in Natural Language Generation}, 
      author={Lorenz Kuhn and Yarin Gal and Sebastian Farquhar},
      year={2023},
      eprint={2302.09664},
      archivePrefix={arXiv},
      primaryClass={cs.CL},
      url={https://arxiv.org/abs/2302.09664}, 
}
\appendix
%% ============================================================

%% ============================================================
%% ============================================================
%%  Appendix
%% ============================================================
%% ============================================================
%%  Appendix
%%  Required packages (add to preamble if not already present):
%%    \usepackage{booktabs}
%%    \usepackage{multirow}
%%    \usepackage{graphicx}      % for \resizebox
%%    \usepackage{enumitem}
%%    \usepackage[skins,breakable]{tcolorbox}
%% ============================================================
\appendix
% ─────────────────────────────────────────────────────────────

\appendix

%% ============================================================
%%  APPENDIX — single-column layout
%% ============================================================
\onecolumn

% ─────────────────────────────────────────────────────────────
\clearpage
\section{Theoretical Foundations}
\label{app:theory}

\vspace{1em}
This appendix clarifies the probability-theoretic foundations underlying each metric, explains why a logically consistent actor must satisfy them, and distinguishes what each metric does and does not measure.

%% ============================================================
\subsection{The Shared Probability Space}
\label{app:probability-space}

Section~\ref{sec:confidence-functions} defined the coherent ideal: a
credence function $c$ whose values are marginals of a single probability
measure $P$ over world-states $\Omega$, with $A_{[x]}(\omega)$ the
answer class correct for $[x]$ in world $\omega$. Throughout this
appendix we abbreviate the correctness event as
\[
E_x := \{\omega \in \Omega : A_{[x]}(\omega) = \hat{y}([x])\},
\]
the event that the model's chosen answer to $x$ is correct, so that
$\bar{c}([x]) = P(E_x)$. Two consequences of this construction carry all
of the structural metrics.

First, \emph{all questions induce events in the same space}. Whether $x$
is a multi-hop question, $x_1$ its first hop, or $x_2$ the conditioned
second hop, the events $E_x$, $E_{x_1}$, $E_{x_2 \mid x_1, y_1^*}$ are
subsets of one $\Omega$, so coherent credences on them are values of one
measure, bound together by the product rule and by monotonicity under
inclusion. Estimating these quantities via separate prompt calls does
not create separate probability spaces; it creates separate
\emph{estimators} of quantities defined over the same underlying space.
Asking an agent ``do you believe it will rain?'' and ``do you believe it
will rain and be cold?'' in different conversations does not exempt the
answers from the conjunction rule.

Second, \emph{the structural metrics test rationalizability rather than
assume rationality}. We make no assumption that the model's epistemic
state has the form above; we ask whether its reported confidences are
consistent with \emph{any} single measure $P$. Reports that are
not admit no coherent probabilistic
interpretation: either the underlying beliefs violate the axioms, or the
estimator does not faithfully transmit them
(Section~\ref{sec:structural-properties}). The Dutch book argument
\cite{ramsey1926} supplies the consequence: an actor whose
confidence reports, read as betting prices, admit no probability
representation can be offered a set of bets that guarantees a loss,
regardless of how the bets are framed or in how many separate
conversations they are placed.

\subsection{Limitations of LLM-Based Semantic Clustering}
\label{app:judge-limitations}

\vspace{0.5em}
We heavily rely on an LLM judge to cluster model responses into semantic equivalence classes, and misclustering is a genuine source of error. Two failure modes are possible: responses expressing the same answer may be split into separate clusters (false negatives), inflating apparent disagreement and artificially lowering SliCK confidence; or responses expressing different answers may be merged (false positives), collapsing distinct classes and artificially raising it. These errors could be propagated in results, however, LLM-based semantic clustering has been shown to be highly reliable on factual QA tasks of the kind we use. Audits of LLM-judge clustering pipelines on short-form factual questions report clustering accuracy of approximately 94\%, with the dominant error type being false negatives on verbose refusal responses rather than misclassification of substantive answers \cite{lamb2026tokentemp}.

%% ============================================================
\subsection{Metric-by-Metric Clarification}
\label{app:metric-clarification}

\vspace{0.5em}
Each metric below follows a common structure: what it measures, and why a consistent actor must satisfy it.

\bigskip

\subsubsection{Normalization}

\paragraph{What it measures.}
Whether the actor's confidence scores over all possible answer classes for a given prompt sum to one.

\paragraph{Why a consistent actor must satisfy it.}
The answer classes $\{[y]\}$ partition the response space: exactly one will be correct. A probability measure over a partition must sum to one by the Kolmogorov axioms. An actor who assigns 0.9 confidence to $[y_1]$ and 0.8 to $[y_2]$ is implicitly claiming the total probability of being correct exceeds 1, a logical impossibility.

\paragraph{Note on SliCK.}
SliCK satisfies normalization by construction: scores are fractions of $k$ rollouts, which sum to one over equivalence classes. We include this metric not as an empirical win for SliCK but as a diagnostic baseline showing how far output-based estimators deviate from a provably achievable property.

\bigskip

\subsubsection{Conjunction Consistency}

\paragraph{What it measures.}
Whether the actor's confidence in a multi-hop question is consistent with its confidences in the constituent sub-questions, as required by the product rule $P(A \cap B) = P(A) \cdot P(B \mid A)$.

\paragraph{Why a consistent actor must satisfy it.}
Let $A = E_{x_1}$ be the event of correctly answering the first hop and $B = E_{x_2 \mid x_1, y_1^*}$ the event of correctly answering the second hop given the first. Correctly answering the full question $x$ is precisely the event $A \cap B$; this is not an approximation but a definition of what multi-hop correctness means. Since $A$, $B$, and $A \cap B$ are all events in the same probability space $\Omega$, the product rule applies directly. An actor who reports $\bar{c}(x) \neq c(x_1, y_1^*) \cdot \bar{c}(x_2 \mid x_1, y_1^*)$ holds beliefs that cannot correspond to any single coherent probability measure.

\paragraph{On separate prompt calls.}
Estimating $\bar{c}(x)$, $c(x_1, y_1^*)$, and $\bar{c}(x_2 \mid x_1, y_1^*)$ via separate prompts does not create separate probability spaces; it creates separate \emph{estimators} of quantities defined over the shared space $\Omega$. Asking a consistent actor the same question in two different conversations does not exempt their answers from the conjunction rule, just as asking a person ``do you believe it will rain?'' and ``do you believe it will rain and be cold?'' separately does not.

\bigskip

\subsubsection{Entailment Monotonicity}

\paragraph{What it measures.}
Whether the actor assigns higher confidence to logically easier questions. If correctly answering $x$ entails correctly answering $x'$, then $E_x \subseteq E_{x'}$ as events in $\Omega$, so monotonicity of probability measures requires $P(E_x) \leq P(E_{x'})$, i.e.\ $\bar{c}(x) \leq \bar{c}(x')$.

\paragraph{Why a consistent actor must satisfy it.}
Set inclusion is preserved by any probability measure: if every world in which you answer $x$ correctly is also a world in which you answer $x'$ correctly, then $P(E_x) \leq P(E_{x'})$ follows immediately. An actor who reports $\bar{c}(x) > \bar{c}(x')$ while accepting $E_x \subseteq E_{x'}$ is asserting a direct contradiction.

\paragraph{The entailment relationship used.}
We use MuSiQue's two-hop structure, where $x'$ is the second hop conditioned on the gold first-hop answer $y_1^*$. Answering the full question $x$ requires answering both hops; answering only the second (given the first) is strictly easier. The entailment $E_x \subseteq E_{x_2 \mid x_1, y_1^*}$ holds by construction of the dataset.

\bigskip

\subsubsection{Prompt Semantic Invariance}

\paragraph{What it measures.}
Whether the actor assigns the same confidence to a question regardless of how it is phrased, when two phrasings are semantically equivalent ($x \simeq x'$).

\paragraph{Why a consistent actor must satisfy it.}
The confidence function $c$ is defined over equivalence classes $[x]$, not surface strings. A consistent actor whose beliefs are about the state of the world must assign $c([x],[y]) = c([x'],[y])$ whenever $[x] = [x']$. Sensitivity to surface form means $\hat{c}$ is tracking features invisible at the semantic level and therefore cannot faithfully represent $c$.

\bigskip

\subsubsection{Generation Semantic Invariance}

\paragraph{What it measures.}
Whether the actor assigns the same confidence to semantically equivalent responses, i.e.\ $\hat{c}(x,y) = \hat{c}(x,y')$ whenever $y \simeq y'$.

\paragraph{Why a consistent actor must satisfy it.}
If $[y]$ is a semantic equivalence class, then $c([x],[y])$ is a function of the class, not the surface string. An estimator that assigns different scores to $y$ and $y'$ within the same class is sensitive to variation that is, by assumption, semantically irrelevant. This cannot represent a coherent probability over semantic outcomes.

\paragraph{Note on SliCK.}
SliCK satisfies this by construction: confidence is assigned at the equivalence-class level as a fraction of rollouts. As with normalization, we include it as a diagnostic baseline rather than as evidence of SliCK's superiority.

%% ============================================================
\subsection{Why Structural Coherence Is Necessary for Usefulness}
\label{app:structural-vs-useful}

\vspace{0.5em}
One might ask whether structural coherence is a necessary condition for usefulness, or merely a theoretical nicety. Our position is that it is necessary: an estimator that violates structural coherence cannot be given a consistent probabilistic interpretation, and therefore cannot be reliably used for downstream decisions that depend on one, such as abstention thresholds, model cascading, or uncertainty-aware aggregation.

An estimator may achieve low RMSCE by saturating near a single value. But if it simultaneously violates normalization, it is not expressing beliefs about the world. It is producing numbers that happen to correlate with accuracy in aggregate while being locally meaningless. Usefulness metrics measure aggregate correlation with ground truth; structural metrics measure whether the scores are interpretable as beliefs at all. Both are necessary.

% ─────────────────────────────────────────────────────────────
\clearpage
\section{Experimental Setup}
\label{app:setup}

\vspace{1em}

\subsection{Models}
\label{app:models}

\vspace{0.5em}
We evaluate nine publicly available instruction-tuned and reasoning models accessed via the OpenRouter API. Table~\ref{tab:models} lists each model with its parameter count, architecture type, and the datasets on which it was evaluated.

\vspace{1em}
\begin{table}[h]
\centering
\caption{Models evaluated in this work. \textbf{SQ} = SimpleQA, \textbf{MQ} = MuSiQue, \textbf{PR} = ParaRel.}
\label{tab:models}
\begin{tabular}{llcll}
\toprule
\textbf{Model} & \textbf{OpenRouter ID} & \textbf{Params} & \textbf{Type} & \textbf{Datasets} \\
\midrule
LLaMA-3.2-1B  & \texttt{meta-llama/llama-3.2-1b-instruct}      & 1B       & Instruct         & SQ, PR \\
LLaMA-3.2-3B  & \texttt{meta-llama/llama-3.2-3b-instruct}      & 3B       & Instruct         & SQ, MQ, PR \\
LLaMA-3-8B    & \texttt{meta-llama/llama-3-8b-instruct}        & 8B       & Instruct         & SQ, MQ, PR \\
LLaMA-3.3-70B & \texttt{meta-llama/llama-3.3-70b-instruct}     & 70B      & Instruct         & SQ, MQ, PR \\
Gemma-3n-E4B  & \texttt{google/gemma-3n-e4b-it}                & $\sim$4B & Instruct         & SQ, PR \\
Gemma-3-27B   & \texttt{google/gemma-3-27b-it}                 & 27B      & Instruct         & SQ, MQ, PR \\
Gemini-2.5-FL & \texttt{google/gemini-2.5-flash-lite}          & n/a      & Instruct         & SQ, MQ, PR \\
Qwen3-14B     & \texttt{qwen/qwen3-14b}                        & 14B      & Instruct         & SQ, MQ, PR \\
DeepSeek-R1   & \texttt{deepseek/deepseek-r1-distill-qwen-32b} & 32B      & Chain-of-thought & SQ, MQ, PR \\
\bottomrule
\end{tabular}
\end{table}

\subsection{Datasets}
\label{app:datasets}

\vspace{0.5em}
We use three benchmarks from the SliCK evaluation suite.

\begin{itemize}
  \setlength\itemsep{0.5em}
  \item \textbf{SimpleQA.} Factual single-hop questions with unambiguous answers. We sample 1,500 questions (seed 42).

  \item \textbf{MuSiQue.} Multi-hop reasoning questions over Wikipedia. Each of the 1,500 sampled questions is decomposed into a full version (all hops) and a hop-1 version (first hop only). The pair structure is used for the Entailment Monotonicity and Conjunction Consistency benchmarks.

  \item \textbf{ParaRel.} Factual probes expressed as semantically equivalent rephrasings of the same underlying relation. We sample 214 conversations covering 35 unique facts, used exclusively for the Prompt Semantic Invariance benchmark.
\end{itemize}

\subsection{Hyperparameters}
\label{app:hyperparams}

\vspace{0.5em}
Table~\ref{tab:hyperparams} lists all hyperparameters used across experiments.

\vspace{1em}
\begin{table}[h]
\centering
\caption{Hyperparameters used across all experiments.}
\label{tab:hyperparams}
\begin{tabular}{lll}
\toprule
\textbf{Stage} & \textbf{Parameter} & \textbf{Value} \\
\midrule
\multirow{4}{*}{Generation}
  & Rollouts per question ($k$) & 16 \\
  & Sampling temperature        & 0.5 \\
  & Max.\ output tokens         & 8{,}192 \\
  & Random seed                 & 42 \\
\midrule
\multirow{3}{*}{Summarization}
  & Judge model        & \texttt{qwen/qwen3-30b-a3b} \\
  & Sampling temperature        & 0.0 \\
  & Max.\ output tokens         & 4{,}096 \\
\midrule
\multirow{3}{*}{Verbal confidence}
  & Judge model        & \texttt{qwen/qwen3-30b-a3b} \\
  & Sampling temperature        & 0.0 \\
  & Max.\ output tokens         & 4{,}096 \\
\midrule
API & Concurrent requests & 15 \\
\bottomrule
\end{tabular}
\end{table}

% ─────────────────────────────────────────────────────────────
\clearpage
\section{Full Benchmark Results}
\label{app:results}

\vspace{1em}
Tables~\ref{tab:results_sq}--\ref{tab:results_mq} report per-model scores across all benchmarks. Both SliCK (frequency-based, \textbf{F}) and Verbal (\textbf{V}) estimators are shown where applicable. Values are accompanied by $\pm$ standard deviation where data permit: per-item s.d.\ for deviation-based metrics (Norm, Gen-SI, CC), mean within-fact s.d.\ for Prompt SI, and binomial s.e.\ for Ent-Mono violation rate. A blank entry denotes that the benchmark was not run for that model or that the model produced no correct answers (precluding AUROC computation).

\vspace{1.5em}

\begin{table}[h]
\centering
\caption{SimpleQA results. AUROC uses the SliCK estimator. Cal = Calibration RMSCE; Norm = Normalization mean absolute deviation; Gen-SI = Generation Semantic Invariance. Subscripts F/V denote SliCK and Verbal estimators. $\downarrow$ lower is better; $\uparrow$ higher is better. Norm\textsubscript{V} can exceed 1 when models report verbal confidence inconsistently (e.g.\ mixing fraction and percentage scales).}
\label{tab:results_sq}
\begin{tabular}{lcccccc}
\toprule
\textbf{Model}
  & \textbf{AUROC}$\uparrow$
  & \textbf{Cal\textsubscript{F}}$\downarrow$
  & \textbf{Cal\textsubscript{V}}$\downarrow$
  & \textbf{Norm\textsubscript{F}}$\downarrow$
  & \textbf{Norm\textsubscript{V}}$\downarrow$
  & \textbf{Gen-SI}$\downarrow$ \\
\midrule
LLaMA-3.2-1B    & \quad      & $0.400$ & $0.074$ & $0.000\,{\pm}\,0.000$ & $0.958\,{\pm}\,0.140$ & $0.000\,{\pm}\,0.000$ \\
LLaMA-3.2-3B    & \quad      & $0.364$ & $0.000$ & $0.000\,{\pm}\,0.000$ & $1.000\,{\pm}\,0.000$ & $0.000\,{\pm}\,0.000$ \\
LLaMA-3-8B      & $0.558$    & $0.271$ & $0.638$ & $0.000\,{\pm}\,0.000$ & $3.565\,{\pm}\,2.956$ & $0.000\,{\pm}\,0.000$ \\
LLaMA-3.3-70B   & $0.591$    & $0.365$ & $0.431$ & $0.000\,{\pm}\,0.000$ & $0.874\,{\pm}\,0.687$ & $0.405\,{\pm}\,0.418$ \\
Gemma-3n-E4B    & $0.572$    & $0.480$ & $0.944$ & $0.000\,{\pm}\,0.000$ & $6.743\,{\pm}\,5.187$ & $0.010\,{\pm}\,0.015$ \\
Gemma-3-27B     & $0.737$    & $0.507$ & $0.944$ & $0.000\,{\pm}\,0.000$ & $4.981\,{\pm}\,4.492$ & $0.000\,{\pm}\,0.000$ \\
Gemini-2.5-FL   & $0.617$    & $0.318$ & $0.817$ & $0.000\,{\pm}\,0.000$ & $5.311\,{\pm}\,4.426$ & $0.150\,{\pm}\,0.333$ \\
Qwen3-14B       & $0.737$    & $0.207$ & $0.806$ & $0.000\,{\pm}\,0.000$ & $7.500\,{\pm}\,4.179$ & $0.000\,{\pm}\,0.000$ \\
DeepSeek-R1-32B & \quad      & \quad   & \quad   & \quad                 & \quad                 & \quad                 \\
\bottomrule
\end{tabular}
\end{table}

\vspace{2em}

\begin{table}[h]
\centering
\caption{ParaRel results. SI = Prompt Semantic Invariance (mean confidence spread across semantically equivalent rephrasings $\pm$ mean within-fact standard deviation). Lower is better.}
\label{tab:results_pr}
\begin{tabular}{lcc}
\toprule
\textbf{Model}
  & \textbf{SI\textsubscript{F}}$\downarrow$
  & \textbf{SI\textsubscript{V}}$\downarrow$ \\
\midrule
LLaMA-3.2-1B    & $0.252\,{\pm}\,0.074$ & $0.207\,{\pm}\,0.076$ \\
LLaMA-3.2-3B    & $0.280\,{\pm}\,0.137$ & $0.545\,{\pm}\,0.257$ \\
LLaMA-3-8B      & $0.498\,{\pm}\,0.182$ & $0.314\,{\pm}\,0.115$ \\
LLaMA-3.3-70B   & $0.625\,{\pm}\,0.207$ & $0.642\,{\pm}\,0.216$ \\
Gemma-3n-E4B    & $0.677\,{\pm}\,0.272$ & $0.123\,{\pm}\,0.049$ \\
Gemma-3-27B     & $0.541\,{\pm}\,0.195$ & $0.170\,{\pm}\,0.056$ \\
Gemini-2.5-FL   & $0.584\,{\pm}\,0.197$ & $0.287\,{\pm}\,0.099$ \\
Qwen3-14B       & $0.522\,{\pm}\,0.185$ & $0.363\,{\pm}\,0.130$ \\
DeepSeek-R1-32B & $0.323\,{\pm}\,0.116$ & $0.311\,{\pm}\,0.121$ \\
\bottomrule
\end{tabular}
\end{table}

\vspace{2em}

\begin{table}[h]
\centering
\caption{MuSiQue results. CC = Conjunction Consistency SliCK (mean absolute deviation $\pm$ per-question s.d.); Ent-Mono = Entailment Monotonicity violation rate $\pm$ binomial s.e. Lower is better for both. A blank entry indicates no MuSiQue data for that model.}
\label{tab:results_mq}
\begin{tabular}{lcc}
\toprule
\textbf{Model}
  & \textbf{CC\textsubscript{F}}$\downarrow$
  & \textbf{Ent-Mono}$\downarrow$ \\
\midrule
LLaMA-3.2-1B    & \quad                 & \quad                 \\
LLaMA-3.2-3B    & $0.320\,{\pm}\,0.344$ & $0.143\,{\pm}\,0.094$ \\
LLaMA-3-8B      & $0.244\,{\pm}\,0.214$ & $0.400\,{\pm}\,0.083$ \\
LLaMA-3.3-70B   & $0.286\,{\pm}\,0.235$ & $0.282\,{\pm}\,0.072$ \\
Gemma-3n-E4B    & \quad                 & \quad                 \\
Gemma-3-27B     & $0.273\,{\pm}\,0.225$ & $0.225\,{\pm}\,0.066$ \\
Gemini-2.5-FL   & $0.247\,{\pm}\,0.240$ & $0.237\,{\pm}\,0.069$ \\
Qwen3-14B       & \quad                 & \quad                 \\
DeepSeek-R1-32B & \quad                 & \quad                 \\
\bottomrule
\end{tabular}
\end{table}

\vspace{2em}

\begin{figure}[h]
\centering
\includegraphics[width=0.85\textwidth]{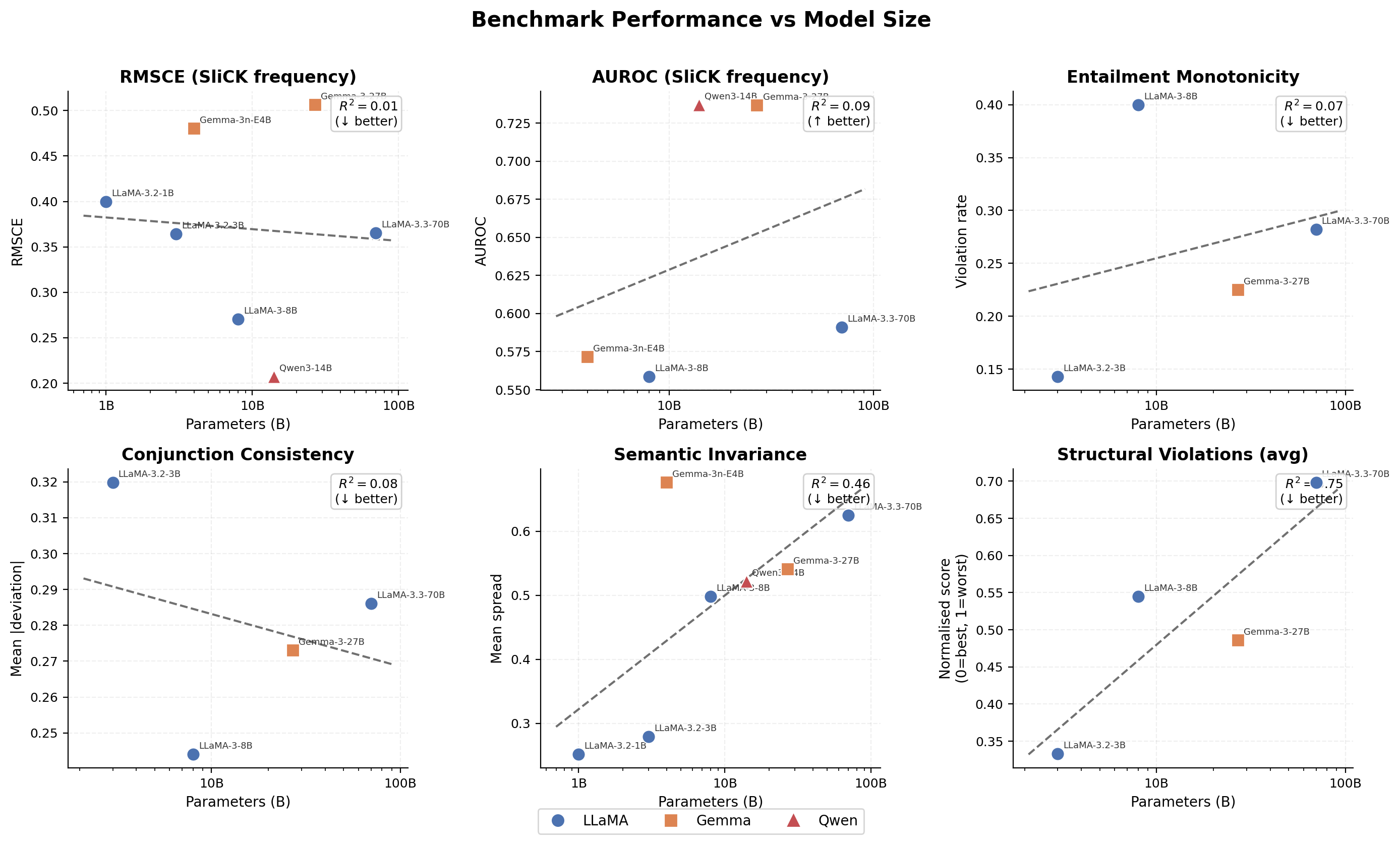}
\caption{\textbf{Scaling laws across metrics.} Structural consistency metrics improve as a function of model scale while most other properties do not follow this trend.}
\label{fig:scaling_laws}
\end{figure}

% ─────────────────────────────────────────────────────────────
\clearpage
\section{Prompts}
\label{app:prompts}

\vspace{1em}

\subsection{Summarization Prompt}
\label{app:prompt_summary}

\vspace{0.5em}
The following prompt is sent to the judge model (\texttt{qwen/qwen3-30b-a3b}) once per rollout to cluster model responses into semantic equivalence classes. Existing cluster labels from prior rollouts are accumulated and passed in the \texttt{\{summaries\}} field so that later rollouts reuse established clusters.

\vspace{1em}
\begin{tcolorbox}[
  colback=gray!5, colframe=gray!35,
  fontupper=\small\ttfamily,
  title={\small Summarization Prompt},
  breakable, left=6pt, right=6pt, top=6pt, bottom=6pt
]
You are an expert data annotation assistant. You will be given:\\[4pt]
1. A question.\\
2. A response to the question.\\
3. Several summaries the final answer could fall under.\\[6pt]
Categorize the final answer as one of the provided summaries, or create a new one if none fits. New summaries must be as short and specific as possible, never a full sentence. For numeric answers, always include the number.\\[6pt]
Do not recompute or verify the answer yourself.\\[6pt]
Question: \{question\}\\[4pt]
Response: \{response\}\\[4pt]
Summaries: \{summaries\}\\[6pt]
Output JSON only:\\
\{"reasoning": <reasoning>, "summary": <summary>\}\\[4pt]
If the model says it does not know, return "Unknown".
\end{tcolorbox}

\vspace{2em}

\subsection{Verbal Confidence Prompt}
\label{app:prompt_verbal}

\vspace{0.5em}
After each rollout is generated, the message below is appended to the conversation and sent back to the \emph{same model under evaluation} (not the judge) at temperature 0 to elicit a scalar confidence score.

\vspace{1em}
\begin{tcolorbox}[
  colback=gray!5, colframe=gray!35,
  fontupper=\small\ttfamily,
  title={\small Verbal Confidence Prompt},
  breakable, left=6pt, right=6pt, top=6pt, bottom=6pt
]
Provide the probability that your guess is correct.\\
Give ONLY the probability, no other words or explanation.\\[6pt]
For example:\\
Probability: <a number between 0.0 and 1.0, nothing else>
\end{tcolorbox}

\vspace{1em}
\noindent
The response is parsed by attempting (in order): a direct \texttt{float()} cast; regex matching for \texttt{Probability:\textbackslash s*([0-9.]+)}; and extraction of the first numeric token. Values outside $[0,1]$ or unparseable responses are recorded as \texttt{None} and excluded from all downstream aggregation.

\vspace{2em}

\subsection{Logit-based Confidence Prompt}
\label{app:prompt_logit}

\vspace{0.5em}
For logit-based confidence, the model is prompted to verify whether a generated response $y$ is true or false for the original question $x$. The prompt below is sent at temperature 0; confidence is computed as the normalized probability of the \texttt{True} token at the first generated position:
\[
\hat{c}(x,y) = \frac{P(\texttt{True})}{P(\texttt{True}) + P(\texttt{False})}.
\]

\vspace{1em}
\begin{tcolorbox}[
  colback=gray!5, colframe=gray!35,
  fontupper=\small\ttfamily,
  title={\small Logit-based Confidence Prompt},
  breakable, left=6pt, right=6pt, top=6pt, bottom=6pt
]
Question: \{question\}\\[4pt]
Proposed answer: \{answer\}\\[6pt]
Is the proposed answer correct? Respond with a single word: True or False.
\end{tcolorbox}

\vspace{1em}
\noindent
The log-probabilities of \texttt{True} and \texttt{False} are extracted from the first token position of the model's response and normalized to sum to one. Responses in which neither token is among the top predicted tokens are excluded from aggregation.

\end{document}